\documentclass[10pt,twocolumn,twoside]{IEEEtran}

\ifCLASSOPTIONcompsoc
  \usepackage[nocompress]{cite}
\else
  \usepackage{cite}
\fi

\usepackage{graphicx}
\usepackage{amsfonts}
\usepackage{amsmath,bm}
\usepackage{url}
\usepackage{endnotes}
\usepackage{algorithm}
\usepackage{algorithmic}
\usepackage{booktabs}
\usepackage{color}
\usepackage{subfigure}
\usepackage{tabularx}
\usepackage{multirow}
\usepackage{verbatim}
\usepackage{pifont}

\usepackage{footmisc}
\usepackage{balance}
\usepackage{hyperref}

\usepackage{xcolor}

\newcommand{\tea}{\textit{SEA~}}
\newcommand{\ie}{\emph{i.e.,~}}
\newcommand{\etal}{\emph{et al.~}}
\newcommand{\eg}{\emph{e.g.,~}}

\newcommand{\cev}[1]{\reflectbox{\ensuremath{\vec{\reflectbox{\ensuremath{#1}}}}}}

\newcommand{\argmax}{\operatornamewithlimits{argmax}}

\newcommand{\best}[1]{\textcolor{blue}{\textbf{#1}}}

\newcommand{\specialcell}[2][l]{%
  \begin{tabular}[#1]{@{}l@{}}#2\end{tabular}}

\newcommand{\PreserveBackslash}[1]{\let\temp=\\#1\let\\=\temp}
\newcolumntype{C}[1]{>{\PreserveBackslash\centering}p{#1}}
\newcolumntype{R}[1]{>{\PreserveBackslash\raggedleft}p{#1}}
\newcolumntype{L}[1]{>{\PreserveBackslash\raggedright}p{#1}}

\newfont{\mycrnotice}{ptmr8t at 7pt}
\newfont{\myconfname}{ptmri8t at 7pt}

\begin{document}
%\thispagestyle{empty} \maketitle \thispagestyle{empty}

%\copyrightyear{2018}
%\acmYear{2020}

%\title{Combining Text Encoders by Multiple Space Learning for Ad-hoc Video Search}
\title{SEA: Sentence Encoder Assembly for Video Retrieval by Textual Queries}

%\author{Xirong Li, Chaoxi Xu, Jiaqi Ji, Fangming Zhou}
%\affiliation{Renmin University of China}

\author{Xirong~Li, \IEEEmembership{Member,~IEEE},~Fangming Zhou,~Chaoxi Xu,~Jiaqi Ji,~Gang Yang
\thanks{Manuscript received August 14, 2020; revised November 05, 2020; accepted November 22, 2020.} 

\thanks{This work was supported in part by NSFC (No. 61672523), Beijing Natural Science Foundation (No. 4202033), the Fundamental Research Funds for the Central Universities and the Research Funds of Renmin University of China (No. 18XNLG19).
The associate editor coordinating the review of this manuscript and approving it for publication was Prof. Raouf Hamzaoui.}

\thanks{The authors are with the Key Lab of Data Engineering and Knowledge Engineering, Renmin University of China, and the AI \& Media Computing Lab, School of Information, Renmin University of China, Beijing 100872, China (e-mail: xirong@ruc.edu.cn) }% <-this % stops a space
}

\markboth{IEEE Transactions on Multimedia, ~Vol.~?, No.~?, ?~2020}%
%\markboth{IEEE Transactions on Image Processing, ~Vol.~?, No.~?, ?~2020}%
%\markboth{IEEE Transactions on Knowledge and Data Engineering,~Vol.~x, No.~x, ~2020}
{Li \MakeLowercase{\textit{et al.}}: SEA: Sentence Encoder Assembly for Video Retrieval by Textual Queries}

\maketitle

%\IEEEtitleabstractindextext{
\begin{abstract}
Retrieving unlabeled videos by textual queries, known as Ad-hoc Video Search (AVS), is a core theme in multimedia data management and retrieval. The success of AVS counts on cross-modal representation learning that encodes both query sentences and videos into common spaces for semantic similarity computation. Inspired by the initial success of previously few works in combining multiple sentence encoders, this paper takes a step forward by developing a new and general method for effectively exploiting diverse sentence encoders. The novelty of the proposed method, which we term \emph{Sentence Encoder Assembly} (SEA), is two-fold. First, different from prior art that use only a single common space, SEA supports text-video matching in multiple encoder-specific common spaces. Such a property prevents the matching from being dominated by a specific encoder that produces an encoding vector much longer than other encoders. Second, in order to explore complementarities among the individual common spaces, we propose multi-space multi-loss learning. As extensive experiments on four benchmarks (MSR-VTT, TRECVID AVS 2016-2019, TGIF and MSVD) show, SEA surpasses the state-of-the-art. In addition, SEA is extremely ease to implement. All this makes SEA an appealing solution for AVS and promising for continuously advancing the task by harvesting new sentence encoders.
\end{abstract}
%}

%\maketitle

%\IEEEdisplaynontitleabstractindextext

\IEEEpeerreviewmaketitle

%\maketitle

%=============================================
%\IEEEraisesectionheading{\section{Introduction} \label{sec:intro}}
\section{Introduction} \label{sec:intro}
%=============================================

\begin{figure*}[thb!]
  \includegraphics[width=\textwidth]{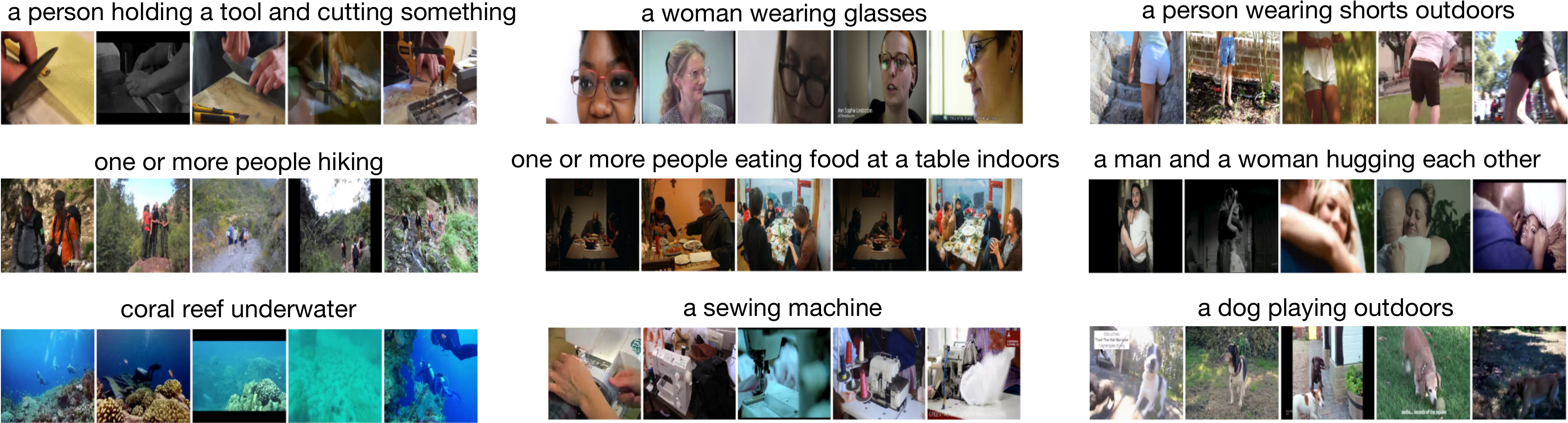}
  \caption{\textbf{Top-5 videos per query sentence retrieved from V3C1~\cite{v3c1}, a large collection of one million \emph{unlabeled} web video clips, by the proposed \tea model}. Queries are from the TRECVID Ad-hoc Video Search benchmarks~\cite{2016trecvidawad,2017trecvidawad,2018trecvidawad,2019trecvidawad}.}  \label{fig:teaser}
\end{figure*}

\IEEEPARstart{V}{ideo} is arguably the most engaging type of digital content in our society. Research related to video content understanding and retrieval is essential for multimedia data management and retrieval. On one hand, common users have been well educated by web search giants such as Google and Baidu to express their information need in textual queries. While on the other hand, there is an increasing amount of videos lacking reliable annotations or even completely unlabeled. This paper targets at the challenging problem of \textit{ad-hoc video search} (AVS), which is to search on many unlabeled videos for user queries expressed exclusively by a phrase or a natural-language sentence and provided on the fly. The complexity of queries varies, ranging from specific objects, \eg ``a sewing machine'', to multi-object events occurred in specific scenes, \eg ``one or more people eating food at a table indoors'', see Fig. \ref{fig:teaser}. A cross-modal similarity model that effectively computes the semantic relevance of the unlabeled videos with respect to a given query is crucial. Also, due to the ad-hoc nature of the query, the model has to be generalizable to handle novel queries unseen when the model is built.

For building such a model, both queries and videos have to be encoded into real-valued vectors via cross-modal representation learning. Earlier efforts struggle to detect semantic concepts from the two modalities and use the detected concepts as an intermediate representation~\cite{lscom,tmm07-hauptmann,ftir09-snoek,icmr16-zsed,icmr2017-certh-avs,icmr17-awad}. Now, it is becoming increasingly evident that learning cross-modal representations in an end-to-end and \textit{concept-free} manner is preferred, as manifested via major benchmarks for the AVS task including TRECVID~\cite{w2vvpp,tv19-alibaba,dong-cvpr19} and MSR-VTT~\cite{mithun-icmr18,liu-bmvc19,luo2020univilm}. 

We concentrate on end-to-end \textit{query representation learning}, an essential component for AVS. Typically, the component is composed of a sentence encoder that vectorizes a textual query into a constant-sized vector and a feed-forward neural network that projects the vector into a common latent space~\cite{miech-iccv19,pan2016jointly,mithun-icmr18,w2vvpp}. 
Varied types of sentence encoders have been investigated in the growing literature.
%, see Fig. \ref{fig:encoder-usage}. 
The vanilla Bag-of-Words (BoW) model is employed by~\cite{habibian-pami17,w2vvpp,dong-cvpr19}, with word2vec (w2v)  in ~\cite{w2vvpp,wray-iccv19}, GRU / bi-GRU in~\cite{mithun-icmr18,w2vvpp,dong-cvpr19,zhang2018cross-modal}, NetVLAD in ~\cite{liu-bmvc19,tv19-alibaba}, and BERT in~\cite{luo2020univilm,mm20-video-retrieval}. While the existing works mainly count on a single sentence encoder, the importance of exploiting multiple sentence encoders for addressing ad-hoc queries has been recognized by few recent works~\cite{w2vvpp,dong-cvpr19,tv19-alibaba}. The W2VV++ model proposed by Li \etal~\cite{w2vvpp} processes a given query by three encoders, \ie BoW, w2v and GRU, in parallel, and then merges the three encoding results by vector concatenation. Dong \etal~\cite{dong-cvpr19} and their follow-up~\cite{tv19-alibaba,icmr20-certh-avs} develop multi-level encoding, where specific sentence encoders are selectively used at distinct levels. Again, vector concatenation is used to combine encodings from the multiple levels. Despite their state-of-the-art performance, we argue that such a concatenation-based method is suboptimal due to the following two reasons. First, the overall encoding could be easily dominated by a specific encoder that produces an encoding vector much longer than the others. For instance, the size of a BoW vector goes up to ten thousand with ease, while encodings of word2vec, GRU or BERT are more compact, with a typical size of a few hundreds. Second, varied encodings by distinct encoders are fed as a whole into the subsequent feed-forward network, meaning the exploration of complementarities among the encoders is limited to a single common space.

%We observe research efforts in three directions that are orthogonal to each other. That is, more training data~\cite{miech-iccv19}, more video features~\cite{mithun-icmr18}, more text encoders~\cite{w2vv}. The third direction has incubated the winning entries for the TRECVID AVS benchmark evaluations in 2018~\cite{w2vvpp} and 2019~\cite{tv19-alibaba}. This work follows the third direction.

In this paper we advance AVS with the following contributions:
\begin{itemize}
	\item We propose \emph{Sentence Encoder Assembly} (SEA), a new and general method for effectively exploiting varied sentence encoders.  \tea bypasses the issues of vector concatenation by learning common spaces per encoder. 
	\item To derive a cross-modal similarity from multiple common spaces, we propose multi-space multi-loss learning, an effective mechanism to explore complementarities among the individual common spaces. 
	\item Our solution surpasses the state-of-the-art on four  benchmarks, \ie MSR-VTT, TRECVID AVS 2016--2019, TGIF and MSVD. Moreover, our solution is easy to implement. With its generality, effectiveness and simplicity, \tea has opened up a promising avenue for harnessing novel sentence encoders for continuous performance improvement of AVS. Code and data are available at \href{https://github.com/li-xirong/sea}{https://github.com/li-xirong/sea}.
\end{itemize}

%known-item search~\cite{jakub-mm19}
%We select W2VV++~\cite{w2vvpp}. The model is used in the winning system in the latest edition of Video Browser Showdown~\cite{jakub-mmm20}. 

\begin{figure*}[tbh!]
\centering
\includegraphics[width=2\columnwidth]{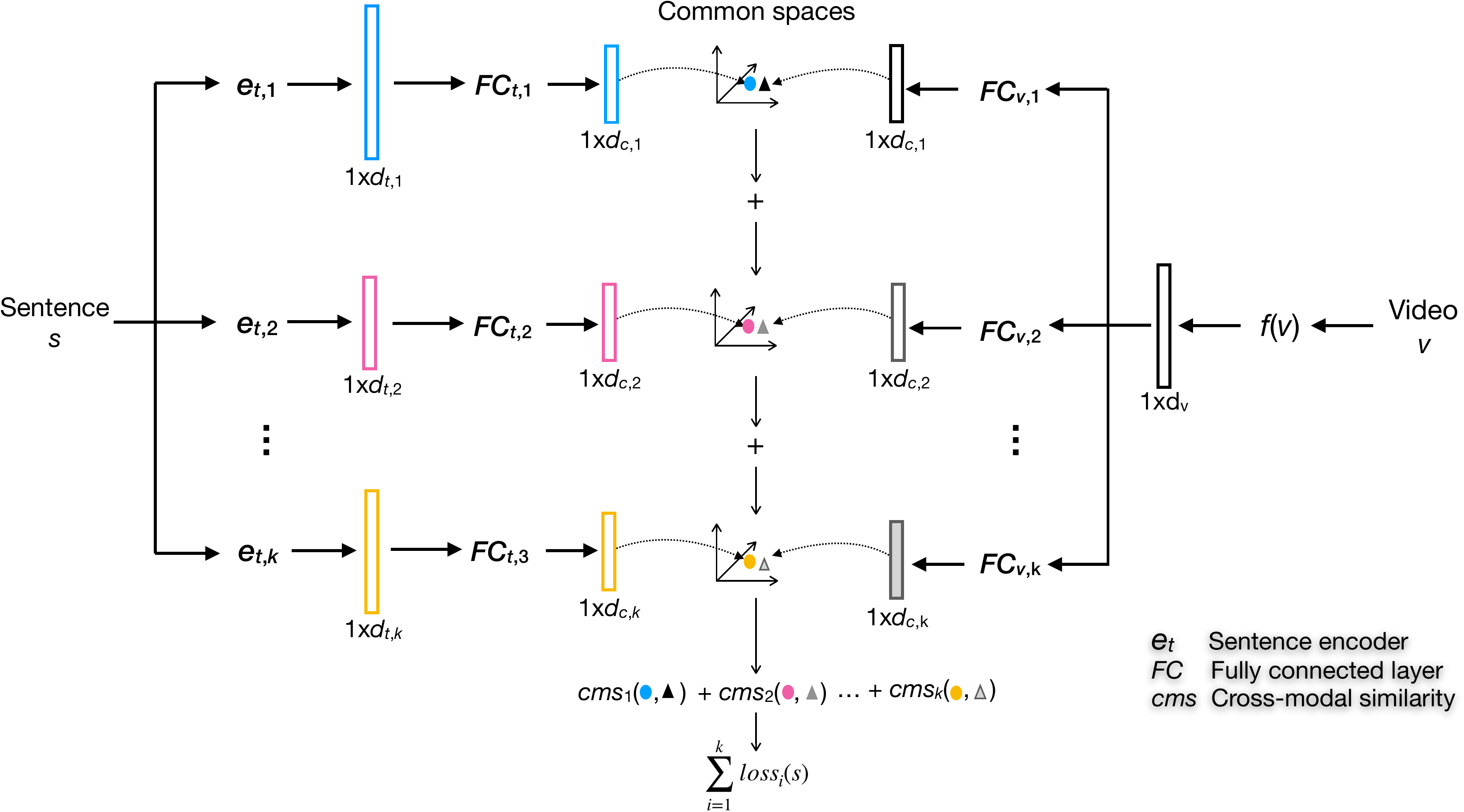}
\caption{\textbf{Proposed Sentence Encoder Assembly (\textit{SEA}) method for exploiting multiple sentence encoders $\{e_{t,1},\ldots,e_{t,k}\}$ for computing cross-modal similarities between a given query sentence $s$ and a specific unlabeled video $v$}. Instead of concatenating the output of the individual sentence encoders as in previous works~\cite{dong-cvpr19,w2vvpp,tv19-alibaba}, our \tea model simultaneously learns $k$ common spaces for the $k$ encoders. Rather than minimizing a single loss computed based on the combined similarity $\sum_{i=1}^k cms_i(s,v)$, \tea is trained to minimize a combine loss $\sum_{i=1}^k loss_i(s)$. Such a multi-space multi-loss learning mechanism is novel and crucial for AVS, meanwhile easy to implement.}
\label{fig:framework}
\end{figure*}

%=============================================
\section{Related Work} \label{sec:related}
%=============================================

%Given that AVS is a rapidly developing topic, we have to selectively discuss the major progress that are measured by TRECVID AVS~\cite{icmr17-awad} and MSR-VTT~\cite{msrvtt}, two leading benchmarks in the field.

Earlier methods for AVS follow a concept-based approach~\cite{ftir09-snoek,icmr16-zsed,tv16-nii,tv17-uva,tv16-certh,tv17-waseda,tv17-vireo}, with both queries and videos represented in a pre-defined concept space. An intrinsic drawback of the concept-based approach is that concepts in use have to be specified in advance, typically according to their occurrence in training data. Such a hand-crafted common space is suboptimal for cross-modal similarity computation~\cite{w2vvpp}. In order to overcome the drawback, end-to-end learning of concept-free and cross-modal representations has been the mainstream~\cite{mithun-icmr18,dong-cvpr19,liu-bmvc19,cvpr20-chen-graph,tmm20-cf-gnn}. As we target at query representation learning, in what follows we discuss recent progress in this direction. 

%We refer to \cite{liu-bmvc19} for combining diverse features for video representation learning, 

The classical Bag-of-Words (BoW) representation is commonly used for its simplicity. In \cite{habibian-pami17}, for instance, a query is first encoded as a BoW vector, and then projected into a latent space through a fully connected layer. However, the BoW encoder has two intrinsic issues. First, it cannot handle semantic relatedness between words. In a BoW feature space, the distance of ``a beagle is running'' to ``a dog is running'' is the same as to ``a person is running'', even though the former pair is visually and semantically more close. Second, it fully ignores word order. To resolve the first issue, Dong \etal \cite{w2vv} employ a pre-trained word2vec model to encode each word in a given query into a dense vector and consequently obtain the query vector by mean pooling over the word-level vectors. Later in Liu \etal ~\cite{liu-bmvc19}, NetVLAD~\cite{netvlad} is adopted to exploit second-order statistics of the word-level vectors.
To overcome the limit of BoW in sequential modeling, varied forms of sequence-aware deep neural networks are investigated. For instance, GRU is used in~\cite{mithun-icmr18}, bi-LSTM in \cite{yu-eccv18}, relational GCN in \cite{cvpr20-chen-graph}, and more recently BERT in \cite{luo2020univilm,mm20-video-retrieval}.

While more advanced sentence encoders are being actively exploited for AVS, it appears to us that no specific encoder is ready to rule them all. We attribute this to the variety and complexity of AVS queries, which can be short phrases or detailed descriptions of multiple-object actions in specific scenes. For the former case, a BoW encoder will suffice, while the latter case requires a complicated encoder to effectively capture fine-grained information. In the context of image/video caption retrieval, Dong \etal \cite{w2vv} make an initial endeavor to combine multiple encoders including BoW, w2v and GRU for query representation. In particular, they concatenate the output of the individual encoders into a lengthy vector. Based on \cite{w2vv}, Li \etal \cite{w2vvpp} develop W2VV++, the winning entry for the TRECVID AVS 2018 evaluation~\cite{tv18-rucmm}. Contemporarily, Dong \etal \cite{dong-cvpr19} propose the Dual Encoding network, wherein three encoders, \ie BoW, bi-GRU and 1-d CNN, are employed to build a multi-level query representation. Follow-ups of \cite{dong-cvpr19}, \eg \cite{tv19-alibaba,tmm20-cf-gnn,icmr20-certh-avs} also leverage multiple encoders, and again merge the output of the encoders in advance to cross-modal representation learning. By contrast, our multi-space learning mechanism makes our model more flexible to harness the complementarities between distinct sentence encoders. Consequently, even with common 2D-CNN features as video representation, our proposed model compares favorably against the state-of-the-art.

%It is worth pointing out 
Note that at a high level, the idea of sentence encoder assembly is similar to the conventional ensemble methods \cite{Bishop06}. However, ensemble learning is a very general idea, typically studied in the context of a classification task. Therefore, a gap naturally exists between the idea itself and putting it to work on AVS. This paper is an initial attempt to bridge the gap. 

\section{Proposed Method}
%=============================================

\subsection{Problem Formalization}

We formalize an ad-hoc video search process as follows. We denote a specific video clip as $v$ and a large collection of $n$ \emph{unlabeled} video clips as $\mathcal{V}=\{v_1,\ldots,v_n\}$. For an ad-hoc query in the form of a sentence $s$, let $cms(s,v)$ be a \textbf{c}ross-\textbf{m}odal \textbf{s}imilarity function that measures the semantic relevance between the query and a specific video. Accordingly, the search process boils down to sorting $\mathcal{V}$ in descending order in terms of $cms(s,v)$ and returning the top ranked items for the given query. The computation of $cms(s,v)$ requires proper embeddings of both $s$ and $v$ into a common cross-modal space. While visual CNNs are prerequisites for video embedding, sentence encoders are required for query embedding. Let $e_t$ be a specific sentence encoder, which encodes the given query into a $d_t$-dimensional real-valued vector, \ie $e_t(s) \in R^{d_t}$.
Having $k$ distinct sentence encoders $\{e_{t,1},\ldots,e_{t_k}\}$ shall give us $k$ vectors of varied dimensions $\{d_{t,1},\ldots,d_{t,k}\}$. We aim for a model that effectively exploit the multiple sentence encoders for computing $cms(s,v)$.

Next, we describe in brief sentence encoders investigated in this work in Section \ref{ssec:tes}, followed by the proposed sentence encoder assembly (SEA) model in Section \ref{ssec:tea}.

\subsection{Sentence Encoders in Use} \label{ssec:tes}

We consider five present-day sentence encoders, \ie Bag-of-Words (BoW), word2vec (w2v), GRU, bi-GRU and BERT. Among them, the first two are unigram, while the others are sequential models. Their main properties are summarized in Table \ref{tab:sent-encoders}.

%including two word-level text encoders, \ie bag-of-words (bow) and word2vec (w2v), and three sentence-level text encoders, \ie GRU, InferSent and BERT.

\begin{table}[tb!]
\normalsize
\renewcommand\arraystretch{1}
\centering
\caption{\textbf{Five sentence encoders used in this paper}. The specific value of the vocabulary size $m$ is dataset-dependent, which is 7,676 for MSR-VTT, 3,981 for TGIF and 2,917 for MSVD. The notation $m$+ means the vocabulary of GRU and bi-GRU is slightly bigger than $m$ due to the inclusion of stopwords and special tokens for sequential modeling.}
\label{tab:sent-encoders}
\scalebox{0.75}{
\begin{tabular}{@{}|l|r|r|l|l|@{}}
\hline
\textbf{Encoder} & \textbf{Vocabulary} & \textbf{Dim.} $d_t$ & \textbf{Training} & \textbf{Prior work} \\
\hline
BoW & $m$ & $m$ & Not trainable & \cite{habibian-pami17,w2vv,w2vvpp,dong-cvpr19} \\
\hline
w2v & 1.7 millions & 500 & \specialcell{pre-trained\footref{w2v-flickr}\\ and fixed} & \cite{w2vv,w2vvpp,wray-iccv19,tv19-alibaba} \\
\hline
GRU & $m$+ & 1,024 & \specialcell{trained from\\ scratch} & \cite{w2vv,w2vvpp,mithun-icmr18} \\
\hline
bi-GRU & $m$+ & 2,048 & \specialcell{trained from\\ scratch} & \cite{dong-cvpr19,tv19-alibaba} \\
%InferSent & 2.0 millions & 4,096 & pre-trained and fixed \\
\hline
BERT & 30,000 & 768 & \specialcell{pre-trained\footref{bert}\\ and fixed} & \cite{luo2020univilm,mm20-video-retrieval}  \\
\hline
\end{tabular}
}
\end{table}

\textbf{1) BoW}. As a classical text encoder, BoW simply quantizes a given sentence $s$ of $l$ words with respect to a pre-specified vocabulary of $m$ words. Let $c(s,j)$ be a function that counts the occurrence of the $j$-th word in the sentence. Accordingly, we have the BoW encoding $e_{BoW}(s)$ as
\begin{equation} \label{eq:bow}
e_{BoW}(s) := (c(s,1), \ldots, c(s,m)).
\end{equation}
Note that an AVS query is relatively short, often containing less than 10 words. Meanwhile, the vocabulary size is much larger, with a typical order of $10^4$. As a consequence, $e_{BoW}(s)$ is a long and sparse vector. 

\textbf{2) w2v}. The w2v model~\cite{w2v} learns to produce word-level dense and semantic vectors by training a two-layer neural network on a large text corpus, with the goal to reconstruct linguistic contexts of words in the training text. As computing the reconstruction loss requires no extra manual annotation, w2v encodes millions of words with ease. We adopt a 500-dimensional w2v model\footnote{\url{https://github.com/danieljf24/w2vv} Note that while \cite{w2vv} performs image-to-text matching experiments on Flickr30k, its w2v model was trained on English tags of 30 million Flickr images, using the skip-gram algorithm.\label{w2v-flickr}} from \cite{w2vv}. 
%, which was trained using the skip-gram algorithm on English tags of 30 million Flickr images. 
We also tried alternatives such as GloVe~\cite{glove}, and found it less effective in our preliminary experiments. Let $w2v(s[i])$ be a lookup function that returns the embedding vector for the $i$-th word of $s$, we obtain w2v based sentence encoding by mean pooling, \ie
\begin{equation} \label{eq:w2v}
e_{w2v}(s) := \frac{1}{l}\sum_{i=1}^l w2v(s[i]).
\end{equation}

\textbf{3) GRU}. The Gated Recurrent Unit (GRU) network~\cite{gru} models the sequential information within a sentence by iteratively generating a sequence of recurrent hidden state vectors $\{\vec{h}_1, \ldots, \vec{h}_l\}$. In particular, the hidden state vector at time-step $i$, $\vec{h}_i$, is jointly determined by the word embedding of the current word $s[i]$ and  $\vec{h}_{i-1}$, the hidden state vector at the previous time-step. Similar to LSTM~\cite{lstm}, the GRU network effectively prevents the vanishing gradient problem by introducing a gating mechanism to modulate the flow of information inside the unit. Meanwhile, as GRU has no separate memory cell, it has a simplified architecture and thus with less parameters to be trained. Following ~\cite{w2vvpp}, we obtain the GRU-based sentence encoding by mean pooling over the hidden vector sequence, \ie
\begin{equation} \label{eq:gru}
e_{gru}(s) := \frac{1}{l}\sum_{i=1}^l \vec{h}_i.
\end{equation}

\textbf{4) bi-GRU}. The bi-directional GRU (bi-GRU) network extends the forward GRU by including a backward GRU that encodes the sequence in a reverse order. Given $\{\cev{h}_1, \ldots, \cev{h}_l\}$ as hidden state vectors of the backward GRU, our bi-GRU based sentence encoding is obtained by 
\begin{equation} \label{eq:gru}
e_{bigru}(s) := \frac{1}{l}\sum_{i=1}^l \vec{h}_i \oplus \cev{h}_i,
\end{equation}
where $\oplus$ denotes vector concatenation. Note that given forward and backward hidden vectors of the same size, $e_{bigru}$ provides a richer representation than $e_{gru}$ at the cost of doubled parameters. Hence, we shall use either $e_{gru}$ or $e_{bigru}$, but not both.

\textbf{5) BERT}. The BERT model, built by stacking a number of $L$ bi-directional Transformer blocks~\cite{bert}, generates word embeddings for a given sentence by progressively passing encodings through the multiple blocks. A Transformer block consists of a self-attention network and a feed-forward network~\cite{transformer}. The self-attention network accepts encodings of individual tokens from the previous Transformer block, weighs their importance to each other by a self-attention mechanism, and accordingly generates new encodings. These encodings are then fed in parallel into the feed-forward network to produce the output encodings of this block. In this work, we adopt the base version of BERT containing $L=12$ blocks, which has been pre-trained on English Wikipedia and book corpora for masked language modeling and next sentence prediction\footnote{\url{https://github.com/google-research/bert}\label{bert}}. We obtain the BERT-based sentence encoding by mean pooling as 
\begin{equation} \label{eq:bert}
e_{bert}(s)=\frac{1}{l} \sum_{i=1}^{l} \mbox{token-emb}(s[i], L-1),
\end{equation}
where $\mbox{token-emb}(s[i], L-1)$ denotes the embedding of the $i$-th word produced by the second-last block. Note that we tried max pooling or using the embedding of the first / last token, and found these alternatives less effective than mean pooling.

%In contrast to recurrent neural networks that process an input sequence of tokens in order, a Transformer block~\cite{transformer} generates encodings for the individual tokens simultaneously. The block consists of a self-attention network and 

With the sentence encoders introduced, we proceed to describe how to effectively combine them in an end-to-end framework.

\subsection{Sentence Encoder Assembly} \label{ssec:tea}

We propose to combine $k$ distinct sentence encoders $\{e_{t,i}|i=1,\ldots,k\}$ in a generic multi-space multi-loss learning framework. 

\textbf{Multiple common spaces}. Our framework consists of $k$ cross-modal matching subnetworks, each corresponding to a specific sentence encoder and learning its own common space. Each subnetwork, indexed by $i$, consists of two fully connected (FC) layers, one on the text side to transform $e_{t,i}(s)$ into a $d_{c,i}$-dimensional vector, and the other on the video side that transforms the video feature vector $f(v)$ into another $d_{c,i}$-dimensional vector. Consequently, the sentence-video semantic relevance, denoted as $cms_i(s,v)$, is computed as the cosine similarity between the two embedding:
\begin{equation} \label{eq:cms_i}
cms_i(s,v) := \mbox{cosine-sim}(\underbrace{FC_{t,i}(e_{t,i}(s))}_{\mbox{text embedding}}, \underbrace{FC_{v,i}(f(v))}_{\mbox{video embedding}}), 
\end{equation}
where $FC_{t,i}$ and $FC_{v,i}$ indicate the two FC layers, each followed by a \textit{tanh} function to increase their learning capacity. We choose the cosine similarity as it is a widely used similarity metric for cross-modal matching~\cite{vsepp,liu-bmvc19,dong-cvpr19,miech-iccv19,sigir20-tce}. We also tried a Euclidean distance based similarity, which is however less effective\footnote{For \tea(\{BoW,w2v\}) trained with the Euclidean distance based similarity, its infAP scores on TV16/17/18/19 are 12.8/18.9/11.8/10.4, clearly lower than the cosine similarity counterpart (15.7/23.4/12.8/16.6).}. 

By simply averaging the similarities computed in the individual common spaces, we have the overall cross-modal similarity as 
\begin{equation} \label{eq:cms}
cms(s,q) := \frac{1}{k} \sum_{i=1}^k cms_i(s,v).
\end{equation}
Note that we do not go for more complicated alternatives, \eg weighing the individual similarities by self-attention mechanisms. 
Rather, we opt for this simple combination strategy, not only for preventing the risk of over-fitting. Such a strategy also encourages the individual common spaces to be good enough to be combined, as they are set to be equally important.

\textbf{Multi-loss learning}. We develop our loss function based on the improved triplet ranking loss (ITRL) by Faghri \etal \cite{vsepp}. While originally proposed for image-text matching, ITRL is now found to be effective for text-video matching~\cite{mithun-icmr18,w2vvpp,dong-cvpr19,liu-bmvc19,tv19-alibaba}. Unlike the classical triplet ranking loss that selects negative training examples by random, ITRL considers the negative that violates the ranking constraint the most (within a mini-batch) and thus deemed to be the most informative for improving the model being trained. Given a training sentence $s$ with $v^+$ as a video relevant w.r.t $s$ and $v^-$ as irrelevant, we express ITRL as 
\begin{equation} \label{eq:single-loss}
\left\{ \begin{array}{ll}
 v^{-*} &= \argmax_{v^- \in batch} (cms(s,v^-) - cms(s,v^+)) \\
 ITRL(s) &=\max(0, \alpha + cms(s,v^{-*}) -cms(s,v^+)), \end{array} \right.
\end{equation}
where $\alpha$ is a positive hyper-parameter concerning the margin.
%, which is set to $0.2$ according to~\cite{vsepp}. 

\begin{figure}[tbh!]
\centering
\includegraphics[width=\columnwidth]{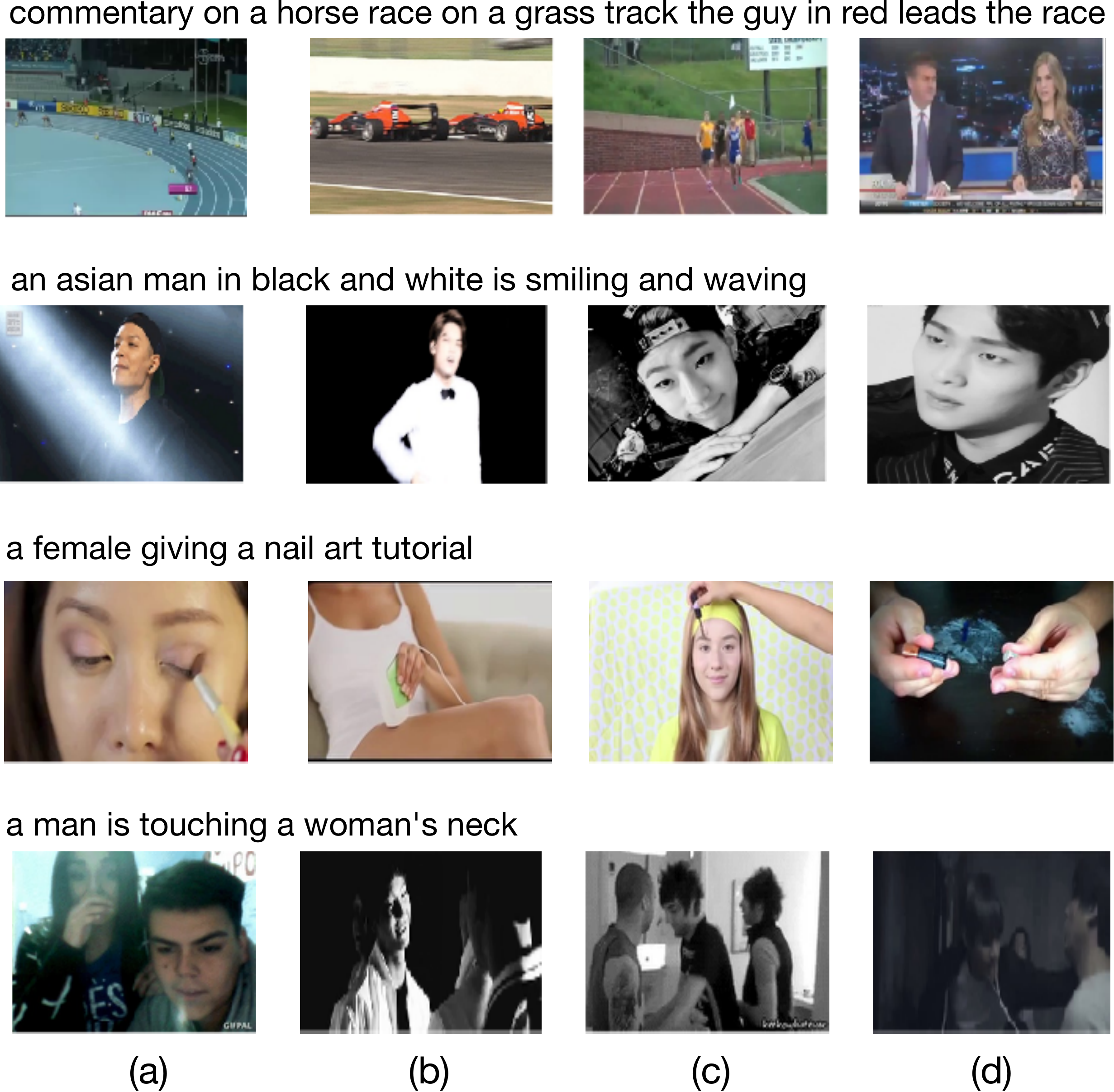}
\caption{\textbf{Examples of hard negative videos automatically selected for specific sentences during training}. The first column (a) is selection based on the combined similarity in a single common space. The other columns indicate selections made based on individual similarities w.r.t (b) $e_{BoW}$, (c) $e_{w2v}$, and (d) $e_{gru}$ within the proposed multi-space and multi-loss framework. Using the combined loss allows the model to be exposed to more diverse hard negatives in a given batch.}
\label{fig:hard-negative}
\end{figure}

We argue that such a single loss is suboptimal for multi-space learning. Given a specific mini-batch, hard negative examples selected in terms of the combined similarity are not necessarily the most effective for learning the individual common spaces. Therefore, we choose to compute $ITRL_i(s)$ per space, and accordingly learn to minimize their combined loss, \ie
\begin{equation}
\sum_{i=1}^k ITRL_i(s).
\end{equation}
In a similar spirit to similarity combination, we again treat all the sub losses equally. As exemplified in Fig. \ref{fig:hard-negative}, the combined loss lets the model be exposed to more diverse hard negatives. We empirically find that compared to the single loss, the combined loss provides around 30\% extra hard negatives per training epoch.

%The total parameters is denoted as $\theta = \theta^{bow} \cup \theta^{w2v} \cup \theta^{gru} \cup \theta^{ms}$. The similarity function is parameterized as $f(s, v; \theta) = \frac{1}{4}(f(s, v; \theta^{bow}) + f(s, v; \theta^{w2v}) + f(s, v; \theta^{gru}) + f(s, v; \theta^{ms}))$.

To sum up, the multi-space strategy provides a more flexible mechanism to exploit complementarities among the distinct sentence encoders. Meanwhile, given a specific mini-batch during training, the multi-loss strategy allows each common space to select its own hard negative example. More flexibility in encoder ensemble and more effectiveness for training together contributes to the superior performance of the proposed SEA method against the state-of-the-art.

%=============================================
\section{Evaluation} \label{sec:exp}
%=============================================

We first conduct experiments on two major benchmarks, MSR-VTT~\cite{msrvtt} and TRECVID AVS~\cite{2016trecvidawad}. While originally developed for video captioning, MSR-VTT has been adopted by recent works for text-based video retrieval~\cite{mithun-icmr18,yu-eccv18,dong-cvpr19,liu-bmvc19,miech-iccv19,cvpr20-chen-graph,luo2020univilm}.
TRECVID AVS is a leading benchmark for ad-hoc video search at large-scale since 2016~\cite{2016trecvidawad,2017trecvidawad,2018trecvidawad,2019trecvidawad}. 
The two benchmarks have their own characteristics. As shown in Table \ref{tab:datasets}, while MSR-VTT has a relatively small amount of 2,990 test videos, it has over 59k query sentences. As for TRECVID, it has a much larger number of test videos, over 335k in the 2016 / 2017 / 2018 editions and over one million in the 2019 edition. Hence, a joint evaluation on the two benchmarks provides a comprehensive assessment of the state-of-the-art. In addition, we report performance on  TGIF~\cite{tgif} and MSVD~\cite{msvd}. % in the Appendix.

%Besides TRECVID, we also  evaluate on the MSR-VTT dataset~\cite{msrvtt}. While originally developed for the purpose of video captioning, this dataset  

\subsection{Experimental Setup}

We first describe experimental setups unique to MSR-VTT and TRECVID, and then introduce common implementations.

\subsubsection{Setup for MSR-VTT}

\begin{table} [tb!]
\renewcommand{\arraystretch}{1}
\caption{\textbf{Datasets used in our evaluation}. For all experiments we train models on the specified training set and use the corresponding validation set for model selection.}
\label{tab:datasets}
\centering
 \scalebox{0.9}{
 \begin{tabular}{@{} |l | l |r |r|r|@{}}
\hline

\textbf{Data split} & \textbf{Data sources} & \textbf{Video clips} & \textbf{Frames} & \textbf{Queries} \\
\hline
\multicolumn{5}{|l|}{\textbf{MSR-VTT experiments:}} \\
\hline
\textit{train set} & \multirow{4}{*}{MSR-VTT~\cite{msrvtt}} & 6,513 & 197,648 & -- \\
\cline{1-1}  \cline{3-5} 
\textit{val. set} &  & 497 & 15,347 & 9,940 \\
\cline{1-1}  \cline{3-5}
\textit{test-full} &  & 2,990 & 92,467 & 59,800 \\
\cline{1-1}  \cline{3-5}
\textit{test-1k}~\cite{yu-eccv18} &  & 1,000 & 30,932 & 1,000 \\

\hline
\hline
\multicolumn{5}{|l|}{\textbf{TRECVID experiments:}} \\
\hline
\multirow{2}{*}{\textit{train set}} & MSR-VTT~\cite{msrvtt}  & 10,000 & 305,462 & --  \\
\cline{2-5}
& TGIF~\cite{tgif}   & 100,855 & 1,045,268 & -- \\
\hline
\textit{val. set} & TV16-VTT-dev~\cite{2016trecvidawad}  & 200 & 5,941 & 200  \\
\hline
\specialcell{\textit{test set} for \\ TV16/17/18} & IACC.3~\cite{2016trecvidawad}  & 335,944 & 3,845,221 & 90 \\
\hline
\specialcell{\textit{test set} for \\TV19} & V3C1~\cite{v3c1} & 1,082,649 & 7,839,450 & 30 \\
\hline
\hline
\multicolumn{5}{|l|}{\textbf{TGIF experiments:}} \\
\hline
\textit{train set} & \multirow{3}{*}{TGIF~\cite{tgif}} & 78,799 & 818,140 & -- \\
\cline{1-1}  \cline{3-5}
\textit{val. set} &  & 10,705 & 110,252 & 10,828 \\
\cline{1-1}  \cline{3-5}
\textit{test set} &  & 11,351 & 116,876 & 34,074 \\
\hline
\hline
\multicolumn{5}{|l|}{\textbf{MSVD experiments:}} \\
\hline
\textit{train set} & \multirow{3}{*}{MSVD~\cite{msvd}} & 1,200 & 23,313 & -- \\
\cline{1-1}  \cline{3-5}
\textit{val. set} &  & 100 & 2,415  & 4,291 \\
\cline{1-1}  \cline{3-5}
\textit{test set} &  & 670 & 15,429  & 27,767 \\
\hline
\end{tabular}
 }% end of scalebox
\end{table}

%\textbf{Data split}. 

We follow the official data split, which divides MSR-VTT into three disjoint subsets used for training, validation and test, respectively. Note that in \cite{yu-eccv18} and its follow-ups~\cite{liu-bmvc19,miech-iccv19,luo2020univilm}, a smaller test set of 1,000 videos randomly sampled from the full test set is used, which we refer to as \textit{test-1k}. 

\textbf{Performance metrics}. Following the previous works, we report \emph{R@k}, $k=1, 5, 10$, the percentage of test queries that have at least one relevant video covered in the top $k$ returned items, and Median rank (\emph{Med r}), the median rank of the first relevant video in the search results. Mean Average Precision (\emph{mAP}) is also reported to assess the overall ranking quality.  

\subsubsection{Setup for TRECVID} \label{sssec:setup-tv}

%\textbf{Data split}. 
We evaluate on the TRECVID AVS testbed from the last four years. The test video collection for TV16 / TV17 / TV18 is IACC.3~\cite{2016trecvidawad}, containing 335,944 web video clips. The test collection for TV19 is V3C1~\cite{v3c-mmm19}, which contains 1,082,649 web video clips, with even more diverse content, no predominant characteristics and low self-similarity~\cite{v3c1}. As no training data is provided by the organizers, we adopt the setup of the winning entry of TV18~\cite{tv18-rucmm}, using MSR-VTT and TGIF~\cite{tgif} for training and the development set of the TV16 video-to-text matching task~\cite{2016trecvidawad} for validation.

\textbf{Performance metric}. The official metric, \ie inferred average precision (infAP)~\cite{infap}, is used.

\subsubsection{Common Implementations} \label{sssec:exp-details}

%\textbf{Video features}.
 We use public feature data\footnote{\url{https://github.com/li-xirong/avs}}, where each video is represented by a 4,096-d feature vector, obtained by using two pre-trained CNNs, \ie ResNet-152 and ResNeXt-101, to extract 2,048-d features from video frames. Frame-level features are concatenated and aggregated to video-level features by mean pooling. We refer to \cite{w2vvpp} for details.

\textbf{Training}. The margin parameter $\alpha$ in the loss is set to $0.2$ according to~\cite{vsepp}. The dimensionality of all the common spaces $d_{c,i}$ is set to 2,048, which achieves a good balance between model performance and model complexity. In fact, our model is highly robust to the choice of the common space dimensionality, see the Appendix.  %We also tried $d_{c,i}$ with a smaller value of 1,024, and found no significant difference. 
We perform SGD based training, with a mini-batch size of 128 and RMSProp as the optimizer. The learning rate is initially set to $10^{-4}$, decayed by a factor of $0.99$ per epoch. Following~\cite{eccv18-weakly}, we half the learning rate if the validation performance does not increase in three consecutive epochs. Early stop occurs when no validation performance increase is achieved in ten consecutive epochs. For each model with specific configurations of sentence encoders, we repeat training three times and pick the version that maximizes the validation performance. All experiments were done with PyTorch (1.2.0)~\cite{pytorch} on an Nvidia GEFORCE GTX 1080Ti GPU.

\begin{table*}[tbh!]
\normalsize
\renewcommand\arraystretch{1}
\centering
\caption{\textbf{Joint evaluation of sentence encoders and their assembly models, \ie W2VV++~\cite{w2vvpp} and the proposed \emph{SEA}, on MSR-VTT and TRECVID}. Numbers are shown in percentages, with best scores shown in \best{bold} font. For a given setup of sentence encoders, relative improvement of \tea over its W2VV++ counterpart is given in parentheses. \tea is consistently better.}
\label{tab:tee_vs_w2vvpp}
\scalebox{0.82}{
\begin{tabular}{@{}|l | l | r | r | r | r | l| r | r | r | r | l|@{}}
\hline
\multirow{2}{*}{\textbf{Sentence encoders}} & \multirow{2}{*}{\textbf{Model}}    & \multicolumn{5}{c|}{\textbf{MSR-VTT} (the full test set)} & \multicolumn{5}{c|}{\textbf{TRECVID} (metric: infAP)}  \\
\cline{3-12}
 & & \textit{R@1} & \textit{R@5} & \textit{R@10} & \textit{Med r} & \textit{mAP} & \textit{TV16} & \textit{TV17} & \textit{TV18} & \textit{TV19} & \textit{MEAN}  \\
\hline
\multirow{2}{*}{\{BoW, w2v\}} & W2VV++ & 10.9 & 29.1 & 39.9 & 19	& 20.2 & 14.4 & 21.8 & 11.1 & 14.3 & 15.4  \\
\cline{2-12}   
 & \tea   & 11.6 & 30.6 & 	41.6 & 17 & 21.3 ($\uparrow$5.4\%) & 15.7 & \best{23.4} & \best{12.8} & 16.6 & \best{17.1} ($\uparrow$11.2\%)\\
\hline
\multirow{2}{*}{\{BoW, w2v, GRU\}} & W2VV++ & 11.1 & 29.6 & 40.5 & 18 & 20.6 & 16.2 & 22.3 & 10.1 & 13.9 & 15.6 	 \\
\cline{2-12}   
 & \tea & 12.2 & 31.9 & 43.1 & 15 & 22.1 ($\uparrow$7.3\%) & 15.0 & \best{23.4} & 12.2 & 16.6 & 16.8 ($\uparrow$7.5\%) \\

%\hline
%\multirow{2}{*}{\{bow, w2v, infersent\}} & W2VV++ & 11.0 & 29.6 & 40.6 & 18 & 20.5 & 14.5 & 22.4 & 11.4 & 14.0 & 62.3 \\
%\cline{2-12}   
% & \tea & 11.4 & 30.5 & 41.6 & 17 & 21.1 ($\uparrow$2.9\%) & 14.6 & 23.2 & 11.9 & \best{16.6} & 66.3 ($\uparrow$6.4\%) \\

\hline
\multirow{2}{*}{\{BoW, w2v, bi-GRU\}} & W2VV++ & 11.3 & 29.9 & 40.6 & 18 & 20.8 & 16.1 & 21.7 & 10.4 & 13.5 & 15.4 \\
\cline{2-12}   
 & \tea & 12.4 & 32.1 & 43.3 & 15 & 22.3 ($\uparrow$7.2\%) & \best{16.4} & 22.8 & 12.5 & \best{16.7} & \best{17.1} ($\uparrow$10.9\%) \\

\hline
\multirow{2}{*}{\{BoW, w2v, BERT\}} & W2VV++ & 12.3	& 31.8 &43.0 & 15 & 22.2 & 15.1 & 22.5 & 10.2 & 12.8 & 15.2  \\
\cline{2-12}   
 & \tea & 12.8 & 33.1  & 44.6 & \best{14} & 23.0 ($\uparrow$3.6\%) & 15.3 & 22.8 & 12.1 & 14.8 & 16.3 ($\uparrow$7.3\%)  \\

\hline
\multirow{2}{*}{\{BoW, w2v, GRU, BERT\}} & W2VV++  & 12.1 & 31.7 & 42.7 & 16 & 22.0 & 14.3 & 19.3 & 9.3 & 10.1 & 13.3  \\
\cline{2-12}   
 & \tea & 13.0 & \best{33.6} & 44.9 & \best{14}	& \best{23.3} ($\uparrow$5.9\%) & 16.0 & 23.1 & 12.1 & 15.4	& 16.7 ($\uparrow$25.7\%)  \\

\hline
\multirow{2}{*}{\{BoW, w2v, bi-GRU, BERT\}} & W2VV++  & 12.0 & 31.3 & 42.3 & 16 & 21.8 & 15.8 & 20.6 & 9.0 & 10.5 & 14.0  \\
\cline{2-12}   
 & \tea & \best{13.1} & 33.4 & \best{45.0} & \best{14}	& \best{23.3} ($\uparrow$6.9\%) & 15.9 & 22.9 & 11.7 & 15.5	& 16.5 ($\uparrow$18.1\%)  \\

\hline 

\end{tabular}
}
\end{table*}

%\subsection{Ablation Study}

%\subsubsection{Joint evaluation of sentence encoders and their assembly models}

\subsection{Experiment 1. Which Sentence Encoders to Use?}

We compare with the state-of-the-art W2VV++~\cite{w2vvpp}, which combines multiple sentence encoders by concatenating their output into a long feature vector and then embeds the concatenated vector into a common space by an FC layer. While originally developed for automated search, W2VV++ has been used with success by Kratochv{\'i}l \etal~\cite{jakub-mmm20} and Loko\'{c} \etal~\cite{mm20-video-retrieval} in the Video Browser Showdown, a leading benchmark for interactive video retrieval~\cite{tmm18-vbs,tmm20-vbs}. For a fair comparison, we use author-provided source code\footnote{\url{https://github.com/li-xirong/w2vvpp}\label{w2vvppcode}} with the same setup as described in Section \ref{sssec:exp-details}. 

The performance of W2VV++ and the proposed \tea model is presented in Table \ref{tab:tee_vs_w2vvpp}. 
For all configurations of sentence encoders, \tea consistently outperforms its W2VV++ counterpart. Specifically, on MSR-VTT our model obtains a relative improvement ranging from 3.6\% to 7.3\% in terms of mAP. While on TRECVID, the relative improvement w.r.t the overall performance ranges from 7.3\% to 25.7\%. The advantage becomes even more clear when four sentence encoders are combined, see the last two rows in Table \ref{tab:tee_vs_w2vvpp}. These results justify the effectiveness of the multi-space mechanism.

As more sentence encoders are included, we observe different phenomenons on the two benchmarks. For MSR-VTT, adding sequential encoders is helpful. Compared to \tea(\{BoW, w2v\}), \tea(\{BoW, w2v, GRU\}) improves mAP from 21.3 to 22.1, while substituting BERT for GRU obtains higher mAP of 23.0. The peak performance, mAP of 23.3, is reached by \tea(\{BoW, w2v, GRU, BERT\}) and \tea(\{BoW, w2v, bi-GRU, BERT\}). By contrast, the inclusion of GRU and BERT has a negative impact on TRECVID. Compared to \tea(\{BoW, w2v\}) which has the best overall infAP of 17.1, adding GRU results in an overall infAP of 16.8, while adding BERT results in a lower value of 16.3. By analyzing the query sentences of the two benchmarks, we find that an MSR-VTT sentence tend to be longer, containing 9.3 words on average, while the corresponding number of TRECVID is 7.1. We attribute this difference to the fact that MSR-VTT was originally meant for video captioning, so its sentences are more detailed. This is furthered confirmed by part-of-speech statistics, where we find that an MSR-VTT query has 3.3 nouns, 1.8 verbs and 0.6 adjective on average, while a TRECVID query has 2.7 nouns, 1.0 verb and 0.4 adjective. TRECVID queries are more keyword-oriented, \eg ``a newspaper'', ``people shopping'', and ``a blond female indoors''. Hence, for answering keyword-oriented queries, \tea(\{BoW, w2v\}) is most suited, while a full setup, \eg \tea(\{BoW, w2v, bi-GRU, BERT\}), is preferred for addressing description-oriented queries.

\begin{table} [tb!]
\renewcommand{\arraystretch}{1}
\caption{\textbf{Evaluating different methods for fusing multiple sentence encoders}, \ie \{BoW, w2v, GRU\}. The most effective method is to train the \tea model with the combined loss.}
\label{tab:fusion}
\centering
 \scalebox{0.95}{
 \begin{tabular}{@{} |l|r|r|r|r|l|@{}}
\hline

\textbf{Fusion method} & \textbf{TV16} & \textbf{TV17} & \textbf{TV18} & \textbf{TV19} & \textbf{MEAN} \\
\hline
W2VV++ & \best{16.2} & 22.3 & 10.1 & 13.9 & 15.6 \\
\hline
Transformed W2VV++ & 13.9 & 20.2 & 10.2 & 13.5 & 14.5 \\
\hline
Model averaging & 14.9 & 21.9 & 11.6 & 15.4 & 16.0  \\
\hline
\tea single loss & 14.7 & 21.8 & 11.2 & 14.7 & 15.6 \\
\hline
\tea combined loss & 15.0 & \best{23.4} & \best{12.2} & \best{16.6} & \best{16.8} \\ 
\hline
\end{tabular}
 }% end of scalebox
\end{table}

\begin{figure*}[tbh!]
\centering
\includegraphics[width=1.7\columnwidth]{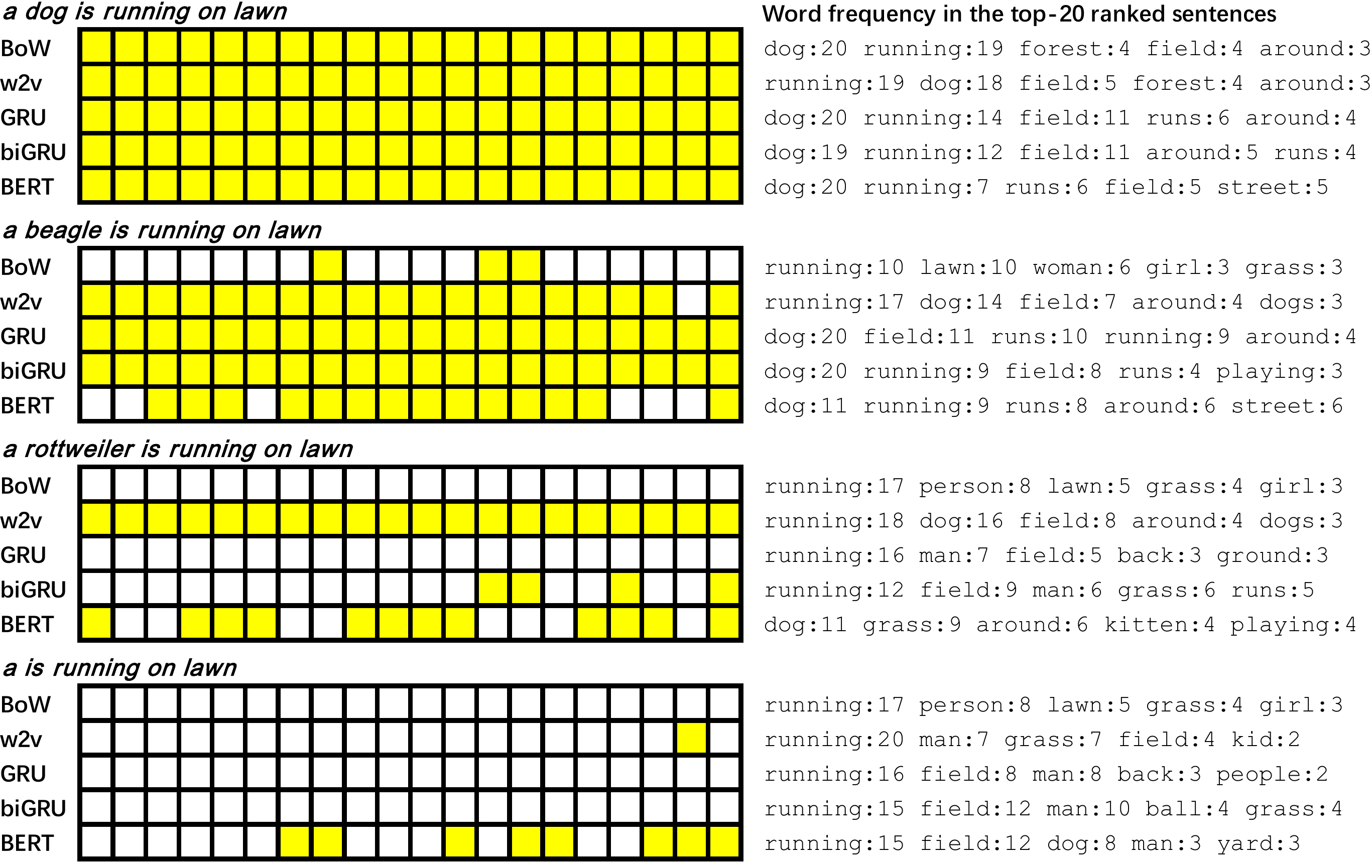}
\caption{\textbf{Visualization of sentence-to-sentence retrieval results}. Given a query sentence, \eg ``a dog is running on lawn'', we retrieve top-20 sentences from MSR-VTT (which has 200k sentences in total), using common spaces learned by cross-modal matching networks with respect to specific sentence encoders. A yellow grid indicates sentences related to dogs. For the last query, we intentionally remove the subject.  Encoders pre-trained on large-scale corpora, \ie w2v and BERT, better handle subjects of low occurrence (`beagle') or zero occurrence (`rottweiler') in the training data. }
\label{fig:sim-sents}
\end{figure*}

We further analyze the complementarity between the distinct sentence encoders by inspecting how they behave when used individually. To that end, we train a cross-modal matching network per encoder on MSR-VTT. To obtain an intuitive understanding of what each network has learned as its common space, we perform sentence-to-sentence retrieval, using all the 200k captions in MSR-VTT as a sentence pool. As w2v and BERT are pre-trained on large-scale corpora with a large vocabulary, they better handle subjects of low occurrence (`beagle') or zero occurrence (`rottweiler’) in the training data, see Fig. \ref{fig:sim-sents}. Interestingly, for the query `a is running on lawn' where we have intentionally remove the subject, BoW returns sentences describing person running, while some of the top-ranked sentences by BERT are still related to dogs. Moreover, we perform a per-query comparison between the matching networks for video retrieval. Among all the 59,800 test queries, the network with BoW is better than the others for 15.2\% of the test queries, while the numbers corresponding to w2v, GRU, bi-GRU and BERT are 14.5\%, 12.9\%, 14.0\% and 19.6\%, respectively. The results clearly show the complementarity between the encoders.

%\subsubsection{Other alternatives for fusing sentence encoders?} 
\subsection{Experiment 2. Other Alternatives for Encoder Assembly?} 

As the output size of the individual sentence encoders ranges from 500 ($e_{w2v}$) up to over 10k ($e_{BoW}$), one might naturally challenge the deficiency of the concatenation operation used by W2VV++. For a more comprehensive comparison, we further implement two more alternatives: \\
$\bullet$ \textit{Transformed W2VV++}. We modify W2VV++ by adding an FC layer after each encoder to transform all encodings into 2,048-d vectors in advance to concatenation. This allows the size of the concatenated vector to be invariant with respect to the encodings. The new variant is also end-to-end trained.  \\
$\bullet$ \textit{Model averaging}. The cross-modal subnetworks w.r.t the individual encoders are trained separately. Cross-modal similarities computed by the subnetworks are equally combined. 

Table \ref{tab:fusion} shows the results of these alternatives on TRECVID. The lower performance of Transformed W2VV++ suggests that adjusting the encodings causes loss in the original information produced by the individual encoders. Model averaging outperforms W2VV++, again suggesting the benefit of using multiple common spaces against a single common space. The result that model averaging is less effective than \tea verifies the necessity of learning multiple common spaces in a unified framework.

%\subsubsection{Combined loss versus single loss}

\subsection{Experiment 3. Combined loss versus Single loss}

\begin{figure}[tbh!]
\centering
\includegraphics[width=\columnwidth]{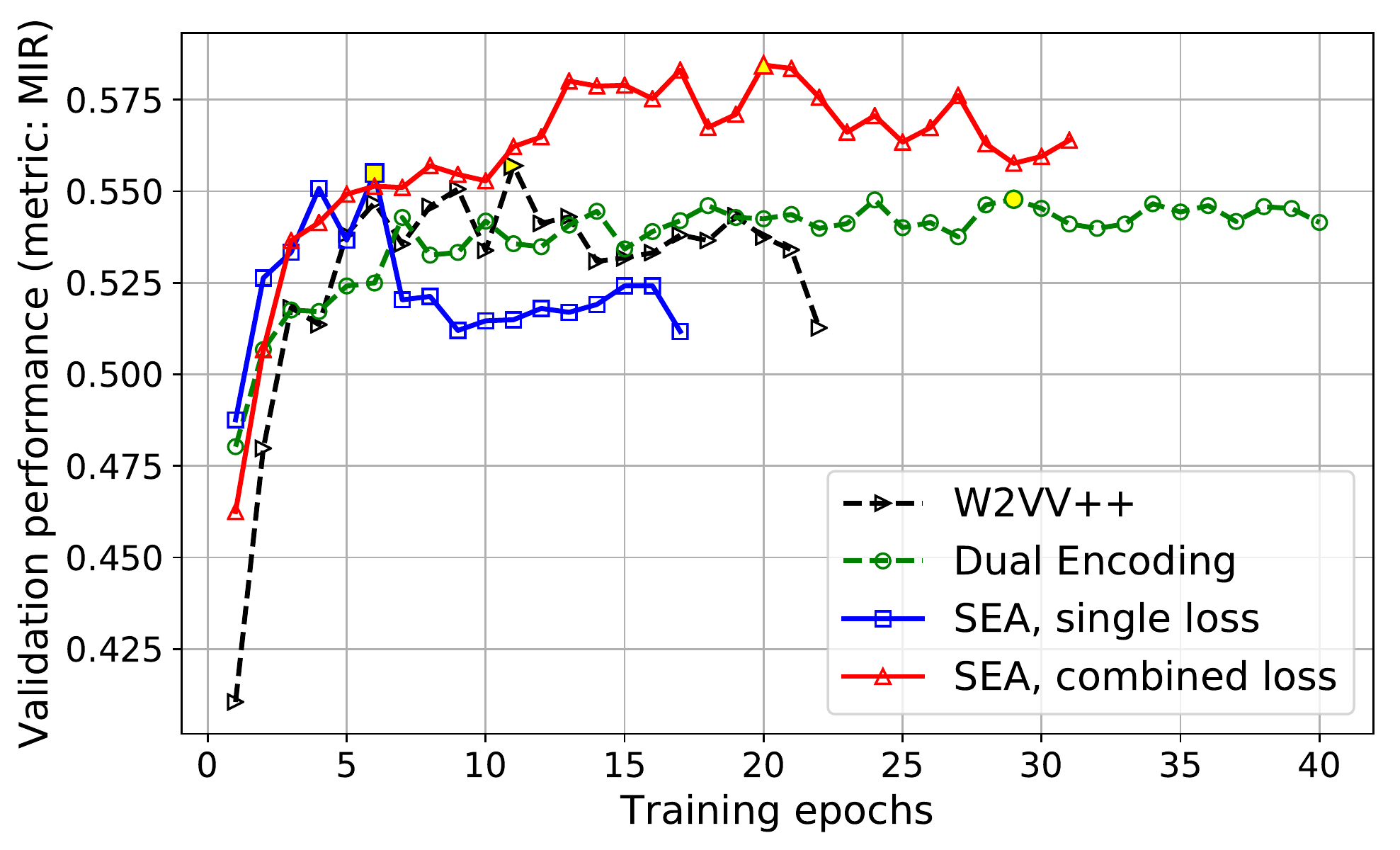}
\caption{\textbf{Learning curves of distinct models in the TRECVID experiments}. Validation is performed after each epoch. The number of epochs a model takes is not pre-specified. After a model reaches its peak performance, as indicated by yellow markers, early stopping occurs in ten epochs.  So the number of training epochs actually took varies among the models. Both W2VV++ and \tea use \{BoW, w2v, GRU\} as their sentence encoders. For training the multi-space network, the combined loss is preferred over the single loss.}
\label{fig:train-curve}
\end{figure}

We have qualitatively illustrate the benefit of the combined loss against the single loss in Fig. \ref{fig:hard-negative}. Now we provide more quantitative evidence. As Table \ref{tab:fusion} shows, \tea trained with the single loss does not outperform W2VV++. We can also observe similar results from the learning curves in Fig. \ref{fig:train-curve}. For both W2VV++ and \emph{SEA}, we use \{BoW, w2v, GRU\} as their sentence encoders. Note that the number of epochs each model takes is not pre-specified. Due to the early stop strategy,  the number of training epochs actually took varies among the models. As shown in Fig. \ref{fig:train-curve}, \tea with the single loss (the blue curve) quickly converged to a suboptimal state. These results proof the importance of the combined loss for training the multi-space network.

\subsection{Comparison to the State-of-the-Art}

\subsubsection{On MSR-VTT}

%\textbf{State-of-the-art}. 
We compare with 11 recent models as follows, which have been evaluated on the \textit{test-1k} set~\cite{yu-eccv18,miech-iccv19,luo2020univilm}, the \emph{full} set~\cite{ijmir19-mithun,cvpr20-chen-graph,tmm20-cf-gnn} or both~\cite{vsepp,liu-bmvc19,dong-cvpr19,w2vvpp,sigir20-tce}. We highlight their choices of sentence encoders: \\
$\bullet$ JSFusion~\cite{yu-eccv18}: Use bi-LSTM as its sentence encoder. \\
$\bullet$ VSE++~\cite{vsepp}: Use GRU as its sentence encoder.\\
$\bullet$ Mithun \etal~\cite{ijmir19-mithun}: Use GRU as its sentence encoder. \\
$\bullet$ Miech \etal~\cite{miech-iccv19}: Use a 1D-CNN as its sentence encoder. \\
$\bullet$ Dual Encoding~\cite{dong-cvpr19}: Hierarchical encoding that combines BoW, bi-GRU and 1D-CNN. \\
$\bullet$ W2VV++~\cite{w2vvpp}: Concatenate encodings of BoW, w2v and GRU \\
$\bullet$ CE~\cite{liu-bmvc19}: Use NetVLAD as its sentence encoder.\\
$\bullet$ TCE~\cite{sigir20-tce}: Use a latent semantic tree for query representation learning. \\
$\bullet$ HGR~\cite{cvpr20-chen-graph}: Encode by  hierarchical semantic graph including three levels of events, actions, entities and relationships across levels. \\
$\bullet$ CF-GNN~\cite{tmm20-cf-gnn}: Graph neural network based search result reranking, with Dual Encoding as its sentence encoder.\\
$\bullet$ UniViLM~\cite{luo2020univilm}: BERT as its sentence encoder. 

Note that all the models were trained on the official training set of MSR-VTT except for~\cite{miech-iccv19,luo2020univilm}, where the authors pre-trained their model on 100 million narrated video clips and then fine-tuned on MSR-VTT.

\begin{table}[tb!]
\normalsize
\renewcommand\arraystretch{1}
\centering
\caption{\textbf{State-of-the-art on MSR-VTT for text-based video retrieval}. Best scores from the cited papers are used, where applicable. On both the \textit{test-1k} set and the \textit{full} test set, our proposed \textit{SEA}(\{BoW, w2v,bi-GRU, BERT\}) is the best.}
\label{tab:sota-msrvtt}
\scalebox{0.8}{
\begin{tabular}{@{}|l |l | r | r | r | r | r|@{}}
\hline
\specialcell{\textbf{Test}\\\textbf{set}} & \textbf{Model} & \textbf{R@1} & \textbf{R@5} & \textbf{R@10} & \specialcell{\textbf{Med}\\\textbf{r}} & \textbf{mAP} \\
\hline
\multirow{4}{*}{\specialcell{\textit{1k}\\\cite{yu-eccv18}}} & JSFusion~\cite{yu-eccv18} & 10.2 & 31.2 & 43.2 & 13 & n.a. \\
\cline{2-7}
%& Miech \etal~\cite{miech-iccv19} & 12.1 & 35.0 &  48.0 & 12 & n.a. \\
& VSE++~\cite{vsepp} & 15.2	& 37.7 & 50.1 & 10 & 26.0 \\
\cline{2-7}
& TCE~\cite{sigir20-tce} & 16.1 & 38.0 & 51.5 & 10 & n.a.\\
\cline{2-7}
& Miech \etal~\cite{miech-iccv19} &14.9 & 40.2 &  52.8 & 9  & n.a. \\
\cline{2-7}
& UniViLM~\cite{luo2020univilm} & 15.4 & 39.5 & 52.3 & 9 & n.a. \\
\cline{2-7}
& Dual Encoding~\cite{dong-cvpr19} & 18.8 & 44.4 & 57.2 & 7 &31.6 \\
\cline{2-7}
& W2VV++~\cite{w2vvpp} & 18.9 & 45.3 & 57.5 & 8 & 31.6 \\
\cline{2-7}
& CE~\cite{liu-bmvc19}  & 20.9 & 48.8 & 62.4 & 6 & n.a. \\
\cline{2-7}
%& \textit{TEE}~(bow,w2v,BERT) & \textbf{22.9} & \textbf{50.4} & \textbf{63.2}	& \textbf{5} & \textbf{35.8} \\
& \tea & \best{23.8} & \best{50.3} & \best{63.8} & \best{5}	& \best{36.6} \\ 

\hline
\hline
\multirow{5}{*}{\textit{Full}} & Mithun \etal~\cite{ijmir19-mithun} & 7.3 & 21.7 & 30.9 & 34 & n.a. \\
\cline{2-7}
& TCE & 7.7 & 22.5 & 32.1 & 30 & n.a. \\
\cline{2-7}
& CF-GNN~\cite{tmm20-cf-gnn} & 8.0 & 23.2 & 32.6 & 31 & 16.0 \\
\cline{2-7}
& VSE++ & 8.7 & 24.3 & 34.1  & 28 & 16.9 \\
\cline{2-7}
& HGR~\cite{cvpr20-chen-graph} & 9.2 & 26.2 & 36.5 & 24 & n.a \\
\cline{2-7}
& CE & 10.0 & 29.0 & 41.2 &	16 & n.a. \\
\cline{2-7}
& Dual Encoding & 11.1 & 29.4 & 40.3 & 19 & 20.5 \\
\cline{2-7}
& W2VV++ & 11.1 & 29.6 & 40.5 &	18	& 20.6 \\
\cline{2-7}
%& \emph{TEE}~(bow,w2v,BERT)  & \textbf{12.8} & \textbf{33.1} & \textbf{44.6} & \textbf{14} & \textbf{23.0} \\ 
& \tea & \best{13.1} & \best{33.4} & \best{45.0} & \best{14}	& \best{23.3} \\
\hline
\end{tabular}
}
\end{table}

\textbf{Results}. Table \ref{tab:sota-msrvtt} shows the performance of the distinct models on the MSR-VTT full test set and \textit{test-1k}. For the ease of comparison, the performance of the baselines is directly cited from the original papers except for W2VV++\footref{w2vvppcode}, VSE++\footnote{\url{https://github.com/fartashf/vsepp}} and Dual Encoding\footnote{\url{https://github.com/danieljf24/dual_encoding}}, which we have re-trained using their public code with the same video feature as used in this work. Among the baselines, CE is the best on \textit{test-1k}, while W2VV++ is the best on the full test set. On both sets, the proposed \tea model is the top performer.
Notice that the good performance of CE is obtained by representing videos with many features including appearance, scene, motion, face, OCR, speech and audio. Given the simplicity of our video feature, the advantage of the new model is clearly justified.

\subsubsection{On TRECVID AVS 2016--2019}

%\textbf{State-of-the-art}. 

\begin{table}[tb!]
\normalsize
\renewcommand\arraystretch{1}
\centering
\caption{\textbf{State-of-the-art on TRECVID AVS}. \tea surpasses the prior art. Late average fusion of two \tea models or \tea(\{BoW,w2v\}) and Dual Encoding boosts the performance.}
\label{tab:sota-tv}
\scalebox{0.7}{
\begin{tabular}{@{}|l | l | l | l | l | c| @{}}
\hline
\textbf{Model} & \textbf{TV16} & \textbf{TV17} & \textbf{TV18} & \textbf{TV19} & \textbf{MEAN} \\
\hline 
\multicolumn{6}{|l|}{Top-3 TRECVID finalists}  \\
\cline{2-6}
Rank 1 & \textcolor{white}{0}5.4\cite{tv16-nii} & 20.6~\cite{tv17-uva} & 12.1~\cite{tv18-rucmm} & 16.3~\cite{tv19-alibaba} & n.a. \\ 
Rank 2 & \textcolor{white}{0}5.1~\cite{tv16-certh} & 15.9~\cite{tv17-waseda} & \textcolor{white}{0}8.7~\cite{tv18-inf} & 16.0~\cite{tv19-rucmm} & n.a.\\
Rank 3 & \textcolor{white}{0}4.0~\cite{tv16-inf} & 12.0~\cite{tv17-vireo} &  \textcolor{white}{0}8.2~\cite{tv18-ntu} & 12.3~\cite{tv19-waseda} & n.a.\\
\hline
VideoStory~\cite{habibian-pami17,tv2017-uva-avs} & \textcolor{white}{0}8.7 & 15.0 & n.a. & n.a. & n.a. \\
\hline
VSE++~\cite{vsepp} & 13.5 & 16.3  & 10.6 & \textcolor{white}{0}9.8  & 12.6\\ 
\hline
W2VV++~\cite{w2vvpp} & 16.2 & 22.3 & 10.1 & 13.9 & 15.6   \\
\hline
Dual Encoding~\cite{dong-cvpr19} & 16.5 & 22.8 & 11.7 & 15.2 & 16.6 \\
\hline
\specialcell{Extended Dual \\ Encoding~\cite{icmr20-certh-avs}} & 15.9 & 24.4 & 12.6 & n.a. & n.a. \\
\hline
\textit{SEA}(\{BoW,w2v\})  & 15.7 & 23.4 & \best{12.8}  & 16.6 & 17.1\\
\hline
\textit{SEA}(\{BoW,w2v,bi-GRU\})  & 16.4 & 22.8 & 12.5  & 16.7 & 17.1\\
\hline
\specialcell{\textit{SEA}(\{BoW,w2v\}) + \\ \textit{SEA}(\{BoW,w2v,bi-GRU\})} & 16.6 & 23.5 & 12.6 & 17.2 & 17.5 \\ 
\hline 
\specialcell{\textit{SEA}(\{BoW,w2v\}) + \\ Dual Encoding} & \best{17.3} & \best{25.0} & \best{12.8} & \best{17.1} & \best{18.1} \\
\hline
\end{tabular}
}
\end{table}

\begin{figure*}[tbh!]
\centering
\includegraphics[width=2\columnwidth]{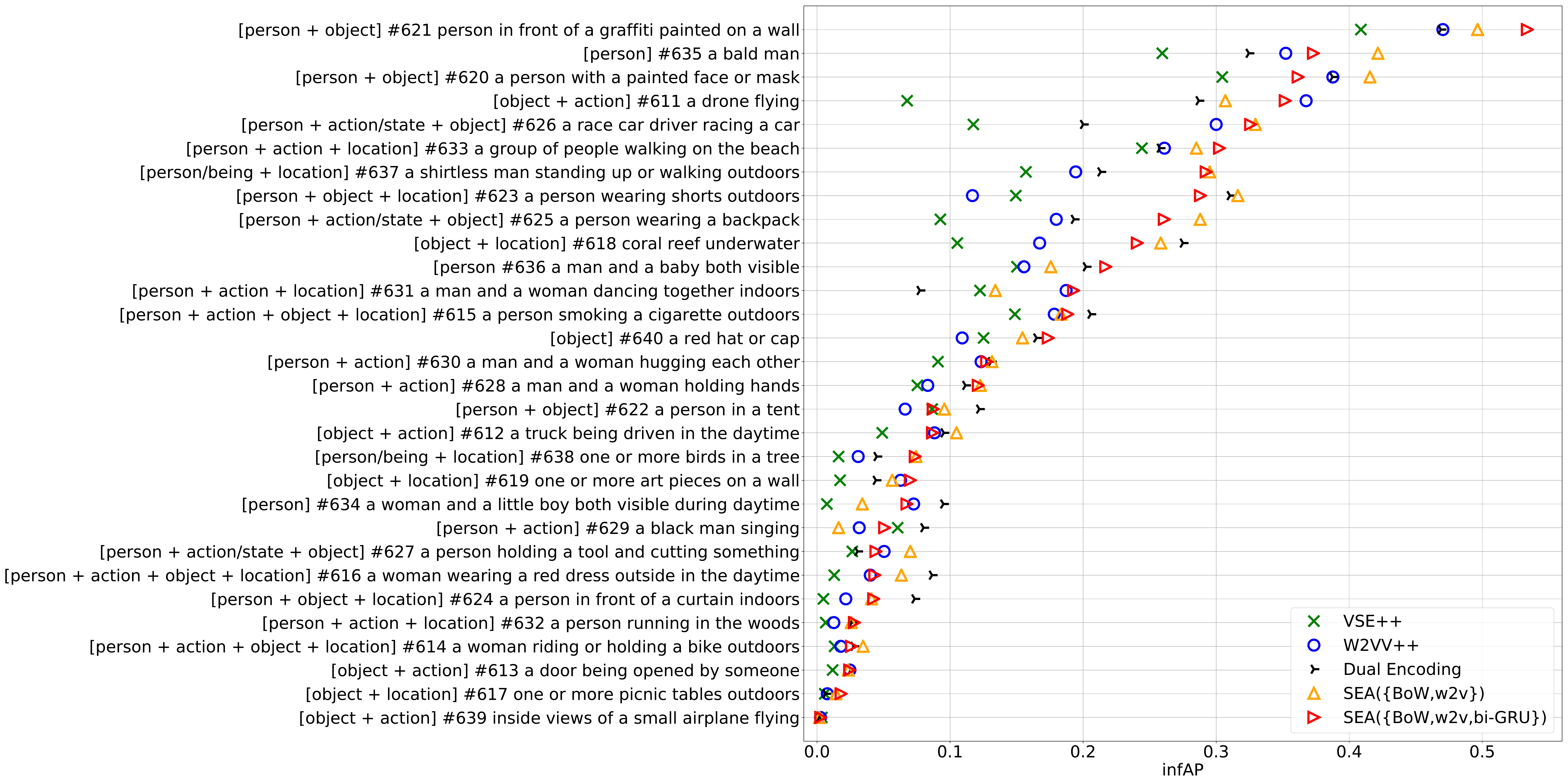}
%\vspace{-4mm}
\caption{\textbf{Performance of varied models in the TRECVID 2019 (TV19) AVS task}, which is to find amidst a set of one million unlabeled videos those relevant with respect to 30 test queries. For result analysis, each query is preceded by the TRECVID-specified query type, \eg [person + object] or [person + action + location], and ID. The queries are sorted in descending order in terms of their infAP scores by \tea(\{BoW,w2v,bi-GRU\}). As the key difference of the varied models is whether multiple sentence encoders are used and how they are combined, the leading performance of the \tea series verifies the effectiveness of the proposed method, namely multi-space network plus multi-loss training. }
\label{fig:tv19}
\end{figure*}

We compare with the top-3 finalist of the TRECVID AVS evaluation each year, which naturally reflects the state-of-the-art. We again compare with W2VV++, VSE++ and Dual Encoding, re-training them using the TRECVID setup as described in Section \ref{sssec:setup-tv}. We also include VideoStory~\cite{habibian-pami17} which uses BoW as its sentence encoder, and Extended Dual Encoding~\cite{icmr20-certh-avs}, a very recent work which makes use of more than one encodings of the visual and textual content and two distinct attention mechanisms.

\textbf{Results}. The performance on the TRECVID test data is shown in Table \ref{tab:sota-tv}. The proposed \tea model surpasses the prior art. While the Extended Dual Encoding network~\cite{icmr20-certh-avs} appears to be on par with the SEA models, \cite{icmr20-certh-avs} has factually used the ground truth of the test set, which shall be unavailable in real applications, to select the best performing models. By contrast, our model selection is performed exclusively based on an independent validation set (see Table \ref{tab:datasets}), and thus more practical. 

 Late average fusion of Dual Encoding and \tea boosts the performance further, see the last row. Note that previous top-performing submissions boost their performance by late (average) fusion of a handful of models~\cite{tv17-uva,tv18-rucmm,tv19-alibaba} or nearly hundred models~\cite{tv19-waseda}. In this context, the capability of \tea to advance the state-of-the-art with a single model is a big advantage for AVS at large-scale.

Fig. \ref{fig:tv19} shows how each model performs on the individual queries from the TV19 task, by searching over the one-million V3C1 collection. Each query is preceded by a TRECVID-specified query type that reflects the query complexity to some extent~\cite{2019trecvidawad}. A query comprised of person, action, object and location tends to be more complex and thus more difficult to address than a query of person. While such a pattern can largely be observed from Fig. \ref{fig:tv19}, exceptions are not uncommon. Consider query \#639, for instance. Although the top-ranked videos show small airplane flying, they are mostly ``external view'', rather than ``inside view'' as required. Such a geometric property has not been effectively captured by the current sentence encoders that are fully data-driven. We consider the \tea model, with its flexibility to harvest new encoders, promising to attack the deficiency.

\subsubsection{On TGIF and MSVD}

For both datasets, we follow the data partition specified by their developers. That is, training / validation / test is 78,799 / 10,705 / 11,351 for TGIF 
%\footnote{The numbers are slightly less than~\cite{tgif} as some  links were invalid when we downloaded the gif files.} 
and 1,200 / 100 / 670 for MSVD. All captions are used. The state-of-the-art following such a setting is HGR~\cite{cvpr20-chen-graph} and CF-GNN~\cite{tmm20-cf-gnn} on TGIF and CF-GNN on MSVD. Therefore, we compare with these two models. Dual Encoding and W2VV++ are also included. 

\textbf{Results}. As shown in Table \ref{tab:sota-tgif} and Table \ref{tab:sota-msvd}, our \tea model is again the best. Given that the amount of the training data in MSVD is substantially less than that of TGIF, the peak performance of \tea on MSVD is reached with less sentence encoders.

\begin{table}[tb!]
\normalsize
\renewcommand\arraystretch{1}
\centering
\caption{\textbf{State-of-the-art on TGIF for text-based video retrieval}.}
\label{tab:sota-tgif}
\scalebox{0.76}{
\begin{tabular}{@{}|l | r | r | r | r | r|@{}}
\hline
\textbf{Model} & \textbf{R@1} & \textbf{R@5} & \textbf{R@10} & \specialcell{\textbf{Med}\\\textbf{r}} & \textbf{mAP} \\
\hline
HGR~\cite{cvpr20-chen-graph} & 4.5 & 12.4 & 17.8 & 160 & n.a. \\
\hline
Dual Encoding~\cite{dong-cvpr19} & 9.1 & 21.3 & 28.6 & 50 & 15.7 \\
\hline
W2VV++~\cite{w2vvpp} & 9.4 & 22.3 & 29.8 & 48 & 16.2 \\
\hline
CF-GNN~\cite{tmm20-cf-gnn} & 10.2 & 23.0 & 30.7& 44& n.a. \\
\hline
\tea(\{BoW,w2v,GRU\}) & 10.2 &23.6 & 31.3 & 41 & 17.2 \\
\hline
\tea(\{BoW,w2v,BERT\})	& 10.7 &24.4 &31.9 &37	& 17.9 \\
\hline
\tea(\{BoW,w2v,GRU,BERT\}) & \best{11.1} & \best{25.2} &32.7 &36&18.4 \\
\hline
\tea(\{BoW,w2v,bi-GRU,BERT\})& \best{11.1} & \best{25.2} & \best{32.8} & \best{35} & \best{18.5} \\ 
\hline
\end{tabular}
}
\end{table}

\begin{table}[tb!]
\normalsize
\renewcommand\arraystretch{1}
\centering
\caption{\textbf{State-of-the-art on MSVD for text-based video retrieval}.}
\label{tab:sota-msvd}
\scalebox{0.76}{
\begin{tabular}{@{}|l | r | r | r | r | r|@{}}
\hline
\textbf{Model} & \textbf{R@1} & \textbf{R@5} & \textbf{R@10} & \specialcell{\textbf{Med}\\\textbf{r}} & \textbf{mAP} \\
\hline
Dual Encoding~\cite{dong-cvpr19} & 20.3	& 46.8 & 59.7 & 6 & 32.9 \\
\hline
CF-GNN~\cite{tmm20-cf-gnn} & 22.8	& 50.9	& 63.6	& 6 & n.a. \\
\hline
W2VV++~\cite{w2vvpp} & 22.4 & 51.6 & 64.8 & 5 & 36.1  \\
\hline
\tea(\{BoW,w2v,GRU\}) & 23.2 &  52.9 &  66.2 &  5 &  37.2  \\
\hline
\tea(\{BoW,w2v,BERT\})	& \best{24.6} & \best{55.0} & \best{67.9} & \best{4} & \best{38.7} \\
\hline
\tea(\{BoW,w2v,GRU,BERT\}) & 24.4& 54.1 & 67.6 & 5& 38.3 \\
\hline
\tea(\{BoW,w2v,bi-GRU,BERT\})& 23.9& 53.9& 67.3& 5& 38.0 \\ 
\hline
\end{tabular}
}
\end{table}

\subsection{Efficiency Analysis}

We report in Table \ref{tab:time_consume} the amount of trainable parameters, training time and inference time of the SEA models with varied setups on MSR-VTT, TGIF and MSVD. Two state-of-the-art methods, \ie Dual Encoding \cite{dong-cvpr19} and W2VV++ \cite{w2vvpp}, are included as well. For a fair comparison, all models use the same size of 2,048 for their common spaces. For all models, the computational cost of video embedding is excluded from the inference time as this step is done once in an offline mode.
Concerning the training time, Dual Encoding is slower than W2VV++ and SEA on MSR-VTT and MSVD, while faster on TGIF. In particular, \tea (\{BoW,w2v,biGRU,BERT\}) requires the longest training time of 4.9 hours on TGIF, as we find that the model needs more training epochs to trigger the early stop mechanism on this dataset.

For each model, its inference time to answer a given query consists of two parts: 1) query embedding that projects the query into a common space (for Dual Encoding and W2VV++) or multiple common spaces (for the \tea models), and 2) ranking that performs cross-modal matching between the query and all videos in a test set and sorting the videos accordingly. The main computational overhead is due to the online inference of the BERT encoder. Still, query embedding can be done within 19 milliseconds. As the cross-modal matching is executed in parallel on GPU, the ranking is extremely fast, costing around one millisecond. The inference time per query is around 20 milliseconds. Hence, our model is sufficiently fast to support real-time interactive video retrieval.

\begin{table*}[tbh!]
  \normalsize
  \renewcommand\arraystretch{1}
  \centering
  \caption{\textbf{Training and inference time of Dual Encoding, W2VV++ and our proposed SEA models on distinct datasets}. The dimensionality of the common space for all models is 2,048. Experiments are done with PyTorch (1.2.0) on an Nvidia 1080Ti GPU.}
  \label{tab:time_consume}
  \scalebox{0.65}{
  \begin{tabular}{@{}|l|r|r|r|r|r|r|r|r|r|r|r|r|@{}}
  \hline
  \multirow{2}[4]{*}{\textbf{Model}} & \multicolumn{4}{c|}{\textbf{MSR-VTT}} & \multicolumn{4}{c|}{\textbf{TGIF}} & \multicolumn{4}{c|}{\textbf{MSVD}} \\

  \cline{2-13}        
  & \specialcell{\#\textit{parameters} \\ (million)} & \specialcell{\textit{training} \\ \textit{time} (hr)} & \specialcell{\textit{query} \\ \textit{embedding} \\ (ms)} & \specialcell{\textit{ranking}\\(ms)} & \specialcell{\#\textit{parameters} \\ (million)} &   \specialcell{\textit{training}\\ \textit{time} (hr)} & \specialcell{\textit{query} \\ \textit{embedding} \\ (ms)} & \specialcell{\textit{ranking}\\(ms)} & \specialcell{\#\textit{parameters} \\ (million)} &  \specialcell{\textit{training} \\ \textit{time} (hr)} & \specialcell{\textit{query} \\\textit{embedding} \\ (ms)} & \specialcell{\textit{ranking}\\(ms)} \\

  \hline
  % Dual Encoding \cite{dong-cvpr19} & 95.9 & 3.4   & 14.8  & 0.6   & 86.5 & 2.5   & 16.6  & 0.6   & 83.7 & 0.7   & 13.7  & 0.4  \\
  % \hline
  % W2VV++ \cite{w2vvpp} & 35.8 & 2.8   & 2.1   & 0.3   & 26.4 & 2.5   & 2.2   & 0.8   & 23.7 & 0.4   & 1.8   & 0.2  \\
  % \hline
  % SEA(BoW,w2v) & 33.5 & 0.9   & 1.2   & 0.5   & 26.0 & 3.1   & 1.0   & 0.9   & 23.8 & 0.1   & 0.9   & 0.3  \\
  % \hline
  % SEA(BoW,w2v,GRU) & 52.6 & 2.7   & 2.4   & 0.6   & 43.2 & 4.4   & 2.4   & 1.0   & 40.5 & 0.2   & 2.2   & 0.4  \\
  % \hline
  % SEA(BoW,w2v,biGRU) & 59.4 & 2.4   & 2.8   & 0.5   & 50.0 & 4.9   & 3.0   & 1.0   & 47.3 & 0.4   & 2.6   & 0.4  \\
  % \hline
  % SEA(BoW,w2v,BERT) & 43.5 & 1.0   & 15.8  & 1.2   & 35.9 & 3.2   & 15.9  & 1.9   & 33.8 & 0.1   & 15.5  & 1.1  \\
  % \hline
  % SEA(BoW,w2v,GRU,BERT) & 62.6 & 2.0   & 17.4  & 1.3   & 53.2 & 4.5   & 17.1  & 1.9   & 50.4 & 0.2   & 17.1  & 1.1  \\
  % \hline
  % SEA(BoW,w2v,biGRU,BERT) & 69.4 & 2.5   & 18.4  & 1.3   & 59.9 & 5.1   & 18.2  & 1.9   & 57.2 & 0.5   & 17.4  & 1.1  \\
  % \hline

  Dual Encoding \cite{dong-cvpr19} & 95.9  & 2.9   & 14.8  & 0.5   & 86.5  & 2.2   & 16.6  & 0.9   & 83.7  & 0.7   & 13.7  & 0.4  \\
  \hline
  W2VV++ \cite{w2vvpp} & 35.8  & 1.2   & 2.1   & 0.5   & 26.4  & 2.4   & 2.2   & 0.9   & 23.7  & 0.3   & 1.8   & 0.4  \\
  \hline
  SEA(BoW,w2v) & 33.5  & 0.9   & 1.2   & 0.6   & 26.0  & 3.1   & 1.0   & 1.0   & 23.8  & 0.1   & 0.9   & 0.6  \\
  \hline
  SEA(BoW,w2v,GRU) & 52.6  & 2.7   & 2.4   & 0.8   & 43.2  & 4.4   & 2.4   & 1.1   & 40.5  & 0.2   & 2.2   & 0.8  \\
  \hline
  SEA(BoW,w2v,bi-GRU) & 59.4  & 2.4   & 2.8   & 0.8   & 50.0  & 4.9   & 3.0   & 1.1   & 47.3  & 0.4   & 2.6   & 0.8  \\
  \hline
  SEA(BoW,w2v,BERT) & 43.5  & 1.0   & 15.8  & 0.8   & 35.9  & 3.6   & 15.9  & 1.1   & 33.8  & 0.2   & 15.5  & 0.8  \\
  \hline
  SEA(BoW,w2v,GRU,BERT) & 62.6  & 2.0   & 17.4  & 0.9   & 53.2  & 4.4   & 17.1  & 1.2   & 50.4  & 0.3   & 17.1  & 1.0  \\
  \hline
  SEA(BoW,w2v,bi-GRU,BERT) & 69.4  & 2.5   & 18.4  & 0.9   & 59.9  & 4.9   & 18.2  & 1.2   & 57.2  & 0.3   & 17.4  & 0.9  \\
  \hline

  \end{tabular}
  }
  \end{table*}

\section{Conclusions} \label{sec:conc}
%=============================================

We have described a method for exploiting diverse sentence encoders for ad-hoc video search. Our experiments show the importance of building a query representation learning network that supports text-video matching in multiple encoder-specific common spaces. Nonetheless, the multi-space network architecture alone is inadequate. In order to effectively utilize complementaries among the individual common spaces,  the network has to be end-to-end trained with a combined loss. On four benchmark datasets including MSR-VTT, TRECVID AVS 2016--2019, TGIF and MSVD,  our proposed \tea model with multi-space multi-loss learning surpasses the prior art.

\appendix

\textbf{The impact of the common space dimensionality}. As shown in Table \ref{tab:cs-dim}, except for using a relatively small value of 256, the dimensionality of the common space has a marginal impact on the performance of the proposed method. We recommend to use 2,048 to strikes a proper balance between performance and model complexity. 

\begin{table}[tb!]
\normalsize
\renewcommand\arraystretch{1}
\centering
\caption{\textbf{The influence of the common space dimensionality $d_{c,i}$ on the model performance}. We evaluate \tea(\{BoW,w2v,GRU\}) on MSR-VTT.}
\label{tab:cs-dim}
\scalebox{0.75}{
\begin{tabular}{@{}|r | r | r | r | r | r|@{}}
\hline
$d_{c,i}$ & \textbf{R@1} & \textbf{R@5} & \textbf{R@10} & \textbf{Med r} & \textbf{mAP} \\
\hline
256 & 11.0 	& 29.5& 40.4 & 18 & 20.4 \\
\hline
512 & 12.0 & 31.5 & 42.7 & 16 & 21.9 \\
\hline
1,024 & 12.1 & 31.6 & 42.9 & \best{15} & 22.0 \\
\hline
2,048 & \best{12.2} & \best{31.9} & \best{43.1} & \best{15} & \best{22.1} \\
\hline
4,096	& \best{12.2} &	31.8 & \best{43.1} & \best{15} & \best{22.1} \\
\hline
8,192  & 12.1 & 31.4 & 42.6 & 16 & 21.8 \\
\hline
\end{tabular}
}
\end{table}

\textbf{The role of pretraining corpus}. As noted in Section \ref{ssec:tes}, we use w2v and BERT which were pre-trained on Flickr tags~\cite{w2vv} and web documents~\cite{bert}, respectively. To investigate if better performance can be obtained by pre-training the two encoders on the same corpus, we have re-trained w2v on Wikipedia dumps and book corpus as used for BERT. We do not try the opposite direction, \ie re-training BERT on the Flickr data, since Flickr tags are not natural-language text and thus unsuited for training BERT. As shown in Table \ref{tab:corpus}, the Flickr version of w2v is slightly better.   

\begin{table}[tb!]
\normalsize
\renewcommand\arraystretch{1}
\centering
\caption{\textbf{Performance of SEA with w2v pre-trained on distinct corpus}. We evaluate \tea(\{BoW,w2v,bi-GRU,BERT\}) on MSR-VTT.}
\label{tab:corpus}
\scalebox{0.75}{
\begin{tabular}{@{}|r | r | r | r | r | r|@{}}
\hline
\textbf{Corpus for w2v} & \textbf{R@1} & \textbf{R@5} & \textbf{R@10} & \textbf{Med r} & \textbf{mAP} \\
\hline
Flickr tags & 13.1 & 33.4 & 45.0 & 14 & 23.3 \\
\hline
Wiki \& book corpus & 13.0 & 33.4 & 44.9 & 14 & 23.3 \\
\hline
\end{tabular}
}
\end{table}

%\textbf{Comparison to the state-of-the-art on TGIF and MSVD}. 

%\newpage 
%\balance

\bibliographystyle{IEEEtran}
\bibliography{msl}

\begin{IEEEbiography}[{\includegraphics[width=1in,clip,keepaspectratio]{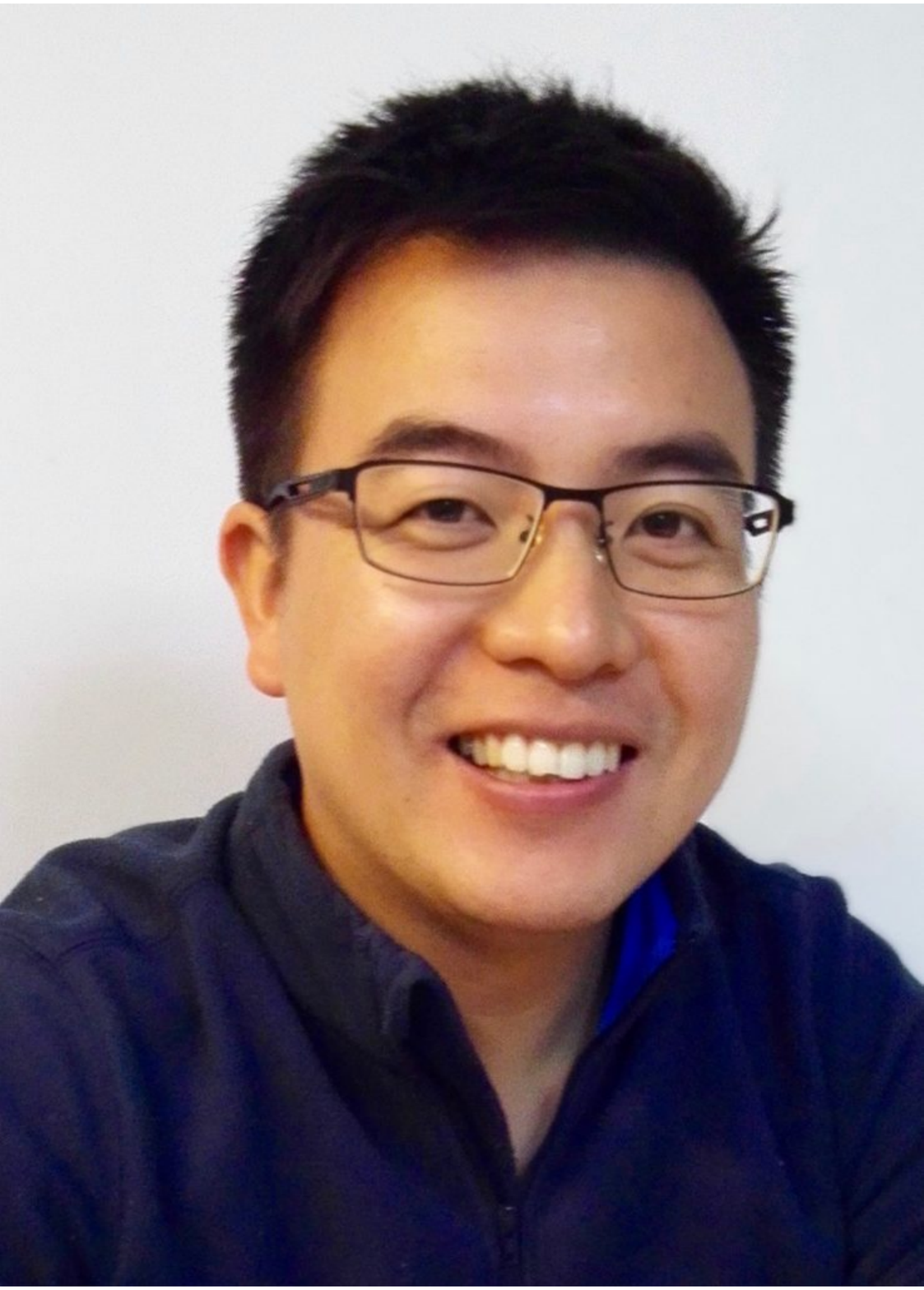}}]
{Xirong Li} received the B.S. and M.E. degrees from Tsinghua University, Beijing, China, in 2005 and 2007, respectively, and the Ph.D. degree from the University of Amsterdam, Amsterdam, The Netherlands, in 2012, all in computer science. He is currently an Associate Professor with the Key Lab of Data Engineering and Knowledge Engineering, Renmin University of China, Beijing, China. His research is multimedia intelligence.

Dr. Li was recipient of the ACMMM 2016 Grand Challenge Award, the ACM SIGMM Best Ph.D. Thesis Award 2013, the IEEE TRANSACTIONS ON MULTIMEDIA Prize Paper Award 2012, and the Best Paper Award of ACM CIVR 2010. He served as program co-chair of Multimedia Modeling 2021. He is associate editor of ACM TOMM and the Multimedia Systems journal.
\end{IEEEbiography}

\begin{IEEEbiography}[{\includegraphics[width=1in,clip,keepaspectratio]{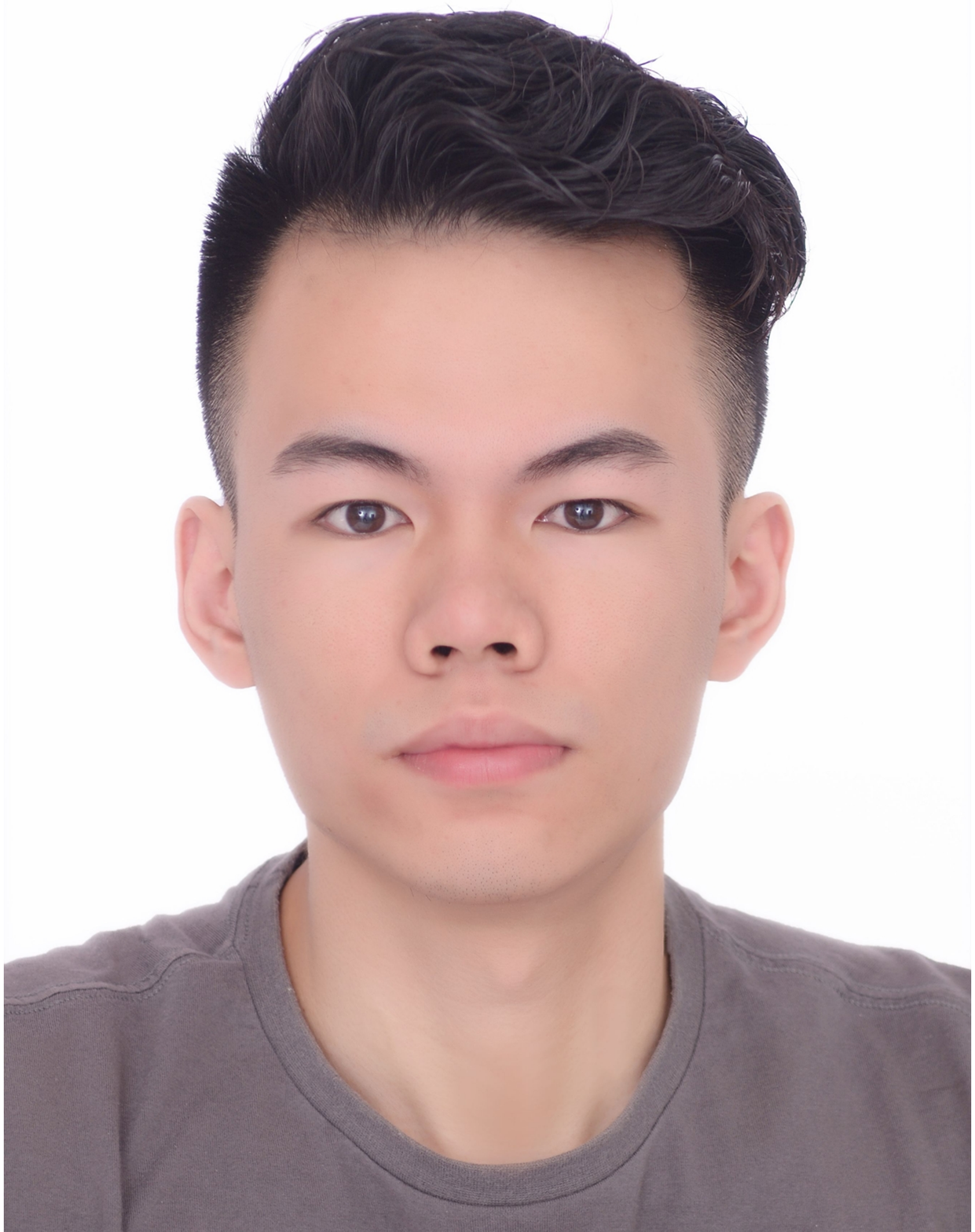}}]
{Fangming Zhou} received his B.S. degree in Nano Materials in Nanjing University of Science and Technology, Nanjing, China in 2019. He is currently a graduate student at School of Information, Renmin University of China, pursuing his master degree on multimedia retrieval.
\end{IEEEbiography}

\begin{IEEEbiography}[{\includegraphics[width=1in,clip,keepaspectratio]{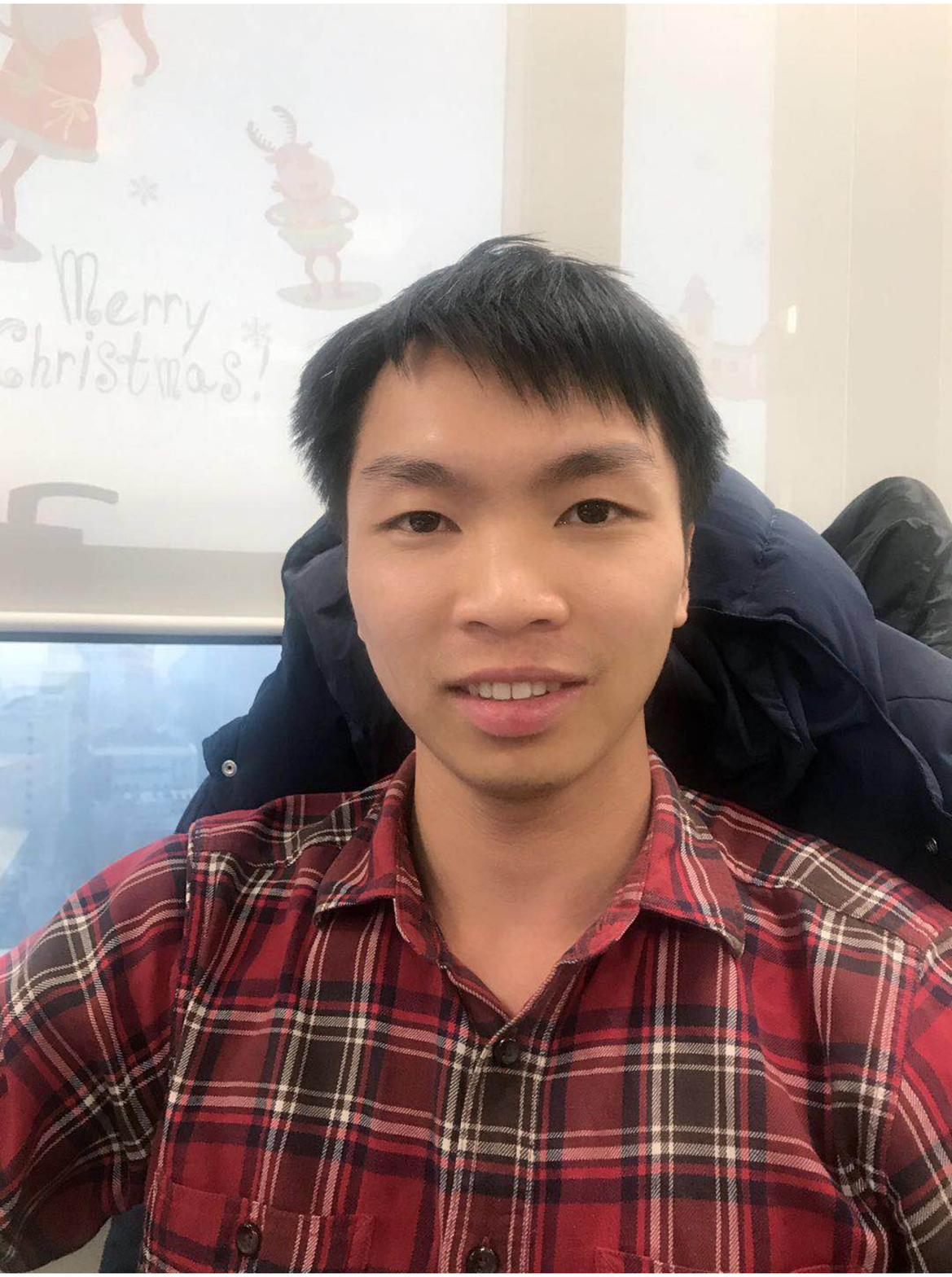}}]
{Chaoxi Xu} received his B.S. and M.E. degrees in Computer Science from Renmin University of China, Beijing, China, in 2017 and 2020, respectively. He is currently an assistant engineer at the Institute of Computing Technology, Chinese Academy of Sciences, Beijing, China.
\end{IEEEbiography}

\begin{IEEEbiography}[{\includegraphics[width=1in,clip,keepaspectratio]{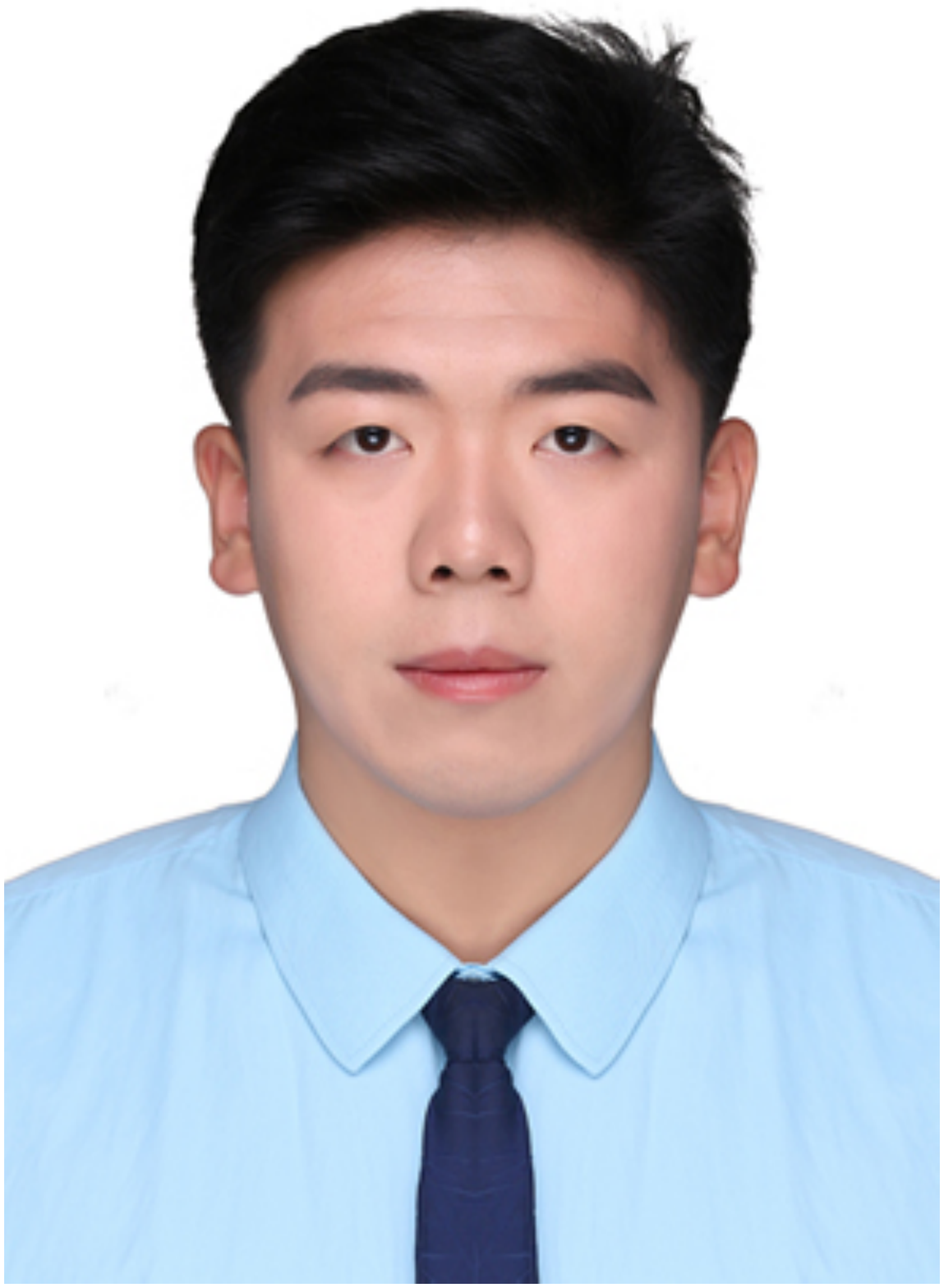}}]
{Jiaqi Ji} received his B.S degree in Software Engineering from Taiyuan University of Technology, Taiyuan, China in 2019. He is currently a graduate student at the School of Information, Renmin  University of china, pursuing his master degree on video retrieval.
\end{IEEEbiography}

\begin{IEEEbiography}[{\includegraphics[width=1in,clip,keepaspectratio]{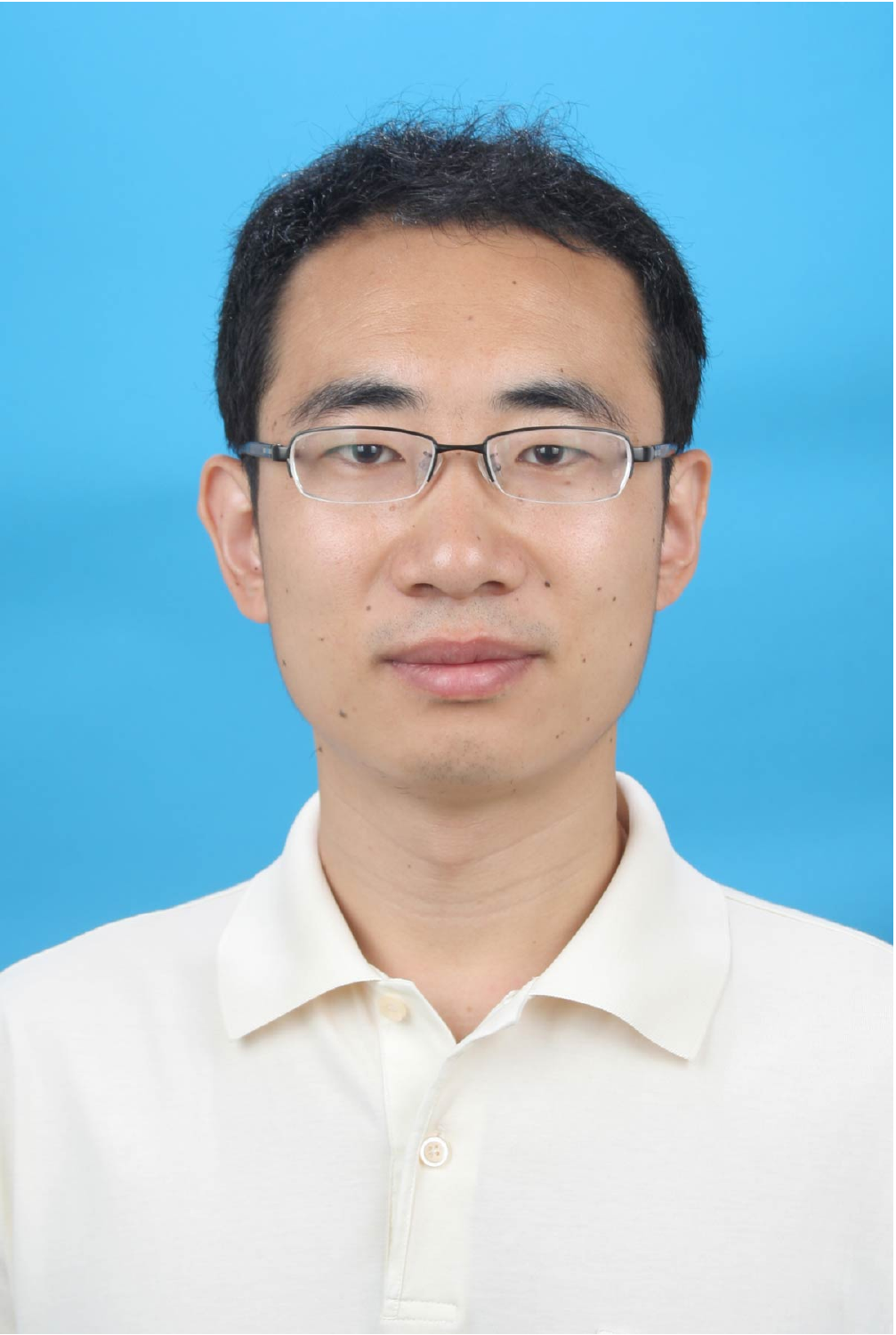}}]
{Gang Yang} received his Ph.D. degree in Innovative Life Science from University of Toyama, Toyama, Japan in 2009. He is currently an Associate Professor at School of Information, Renmin University of China, Beijing, China. His research interests include computational intelligence, multimedia computing and machine learning.
\end{IEEEbiography}

\end{document}